\documentclass{latex_template/vgtc}                          % final (conference style)
\ifpdf%                                % if we use pdflatex
  \pdfoutput=1\relax                   % create PDFs from pdfLaTeX
  \usepackage{graphicx}                % allow us to embed graphics files
  \DeclareGraphicsExtensions{.pdf,.png,.jpg,.jpeg} % for pdflatex we expect .pdf, .png, or .jpg files
\else%                                 % else we use pure latex
  \usepackage{graphicx}                % allow us to embed graphics files
  \DeclareGraphicsExtensions{.eps}     % for pure latex we expect eps files
\fi%

\usepackage[whole]{bxcjkjatype}
\usepackage{url}

\graphicspath{{figures/}{pictures/}{images/}{./}{template_2019/pictures/}{fig/}} % where to search for the images

\usepackage{microtype}                 % use micro-typography (slightly more compact, better to read)
\PassOptionsToPackage{warn}{textcomp}  % to address font issues with \textrightarrow
\usepackage{textcomp}                  % use better special symbols
\usepackage{mathptmx}                  % use matching math font
\usepackage{times}                     % we use Times as the main font
\usepackage{cite}                      % needed to automatically sort the references
\usepackage{tabu}                      % only used for the table example
\usepackage{booktabs}                  % only used for the table example
\usepackage{multirow}

\onlineid{1060}

\vgtccategory{Research}

\vgtcinsertpkg

\title{Retargetable AR: Context-aware Augmented Reality in Indoor Scenes based on 3D Scene Graph}

\author{Tomu Tahara\thanks{\tt\small e-mail:\{tomu.tahara, takashi.seno, gaku.narita, tomoya.ishikawa\}@sony.com}\\ %
\and Takashi Seno\\ %
\and Gaku Narita\\ %
\and Tomoya Ishikawa\\ %
    }
\affiliation{\scriptsize R\&D Center \\ Sony Corporation \vspace{-0.5\baselineskip}}

\teaser{
 \includegraphics[width=\columnwidth]{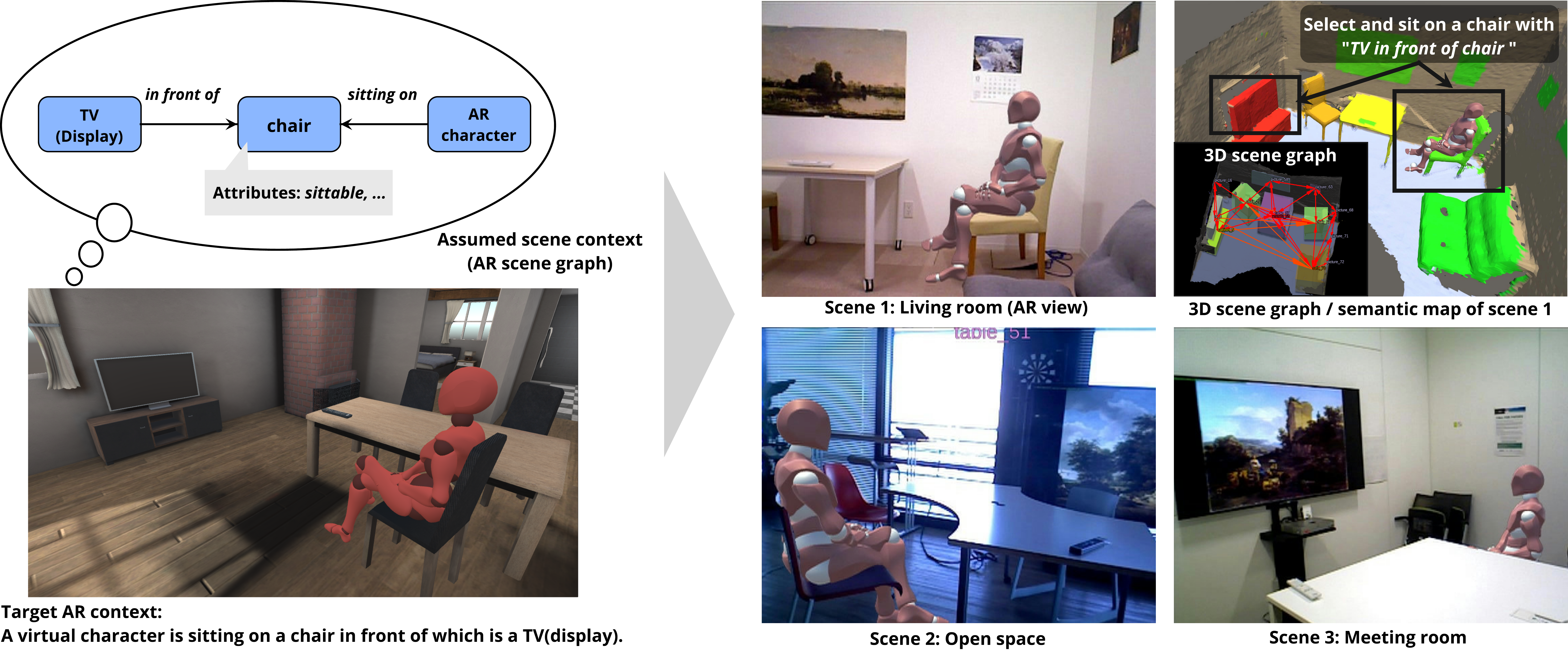}
 \centering
 \caption{\textbf{\textit{Retargetable AR}}. The context of an AR scene in which "A virtual character is sitting on a chair in front of which is a TV" is expressed as an abstract AR scene graph based on the relationships among objects (left). 
 The framework can retarget the AR scene to various real scenes by comparing the AR scene graph with 3D scene graphs constructed in each of the scenes  (right). 
 }
 \label{fig:teaser_overview}
}

\abstract{
  In this paper, we present \textbf{\textit{Retargetable AR}}---a novel AR framework that yields an AR experience that is aware of scene contexts set in various real environments, achieving natural interaction between the virtual and real worlds.
  To this end, we characterize scene contexts with relationships among objects in 3D space, not with coordinates transformations. A context assumed by an AR content and a context formed by a real environment where users experience AR are represented as abstract graph representations, i.e. scene graphs.
  From RGB-D streams, our framework generates a volumetric map in which geometric and semantic information of a scene are integrated.
  Moreover, using the semantic map, we abstract scene objects as oriented bounding boxes and estimate their orientations.
  With such a scene representation, our framework constructs, in an online fashion, a 3D scene graph characterizing the context of a real environment for AR.
  The correspondence between the constructed graph and an AR scene graph denoting the context of AR content provides a semantically registered content arrangement, which facilitates natural interaction between the virtual and real worlds.
  We performed extensive evaluations on our prototype system through quantitative evaluation of the performance of the oriented bounding box estimation, subjective evaluation of the AR content arrangement based on constructed 3D scene graphs, and an online AR demonstration.
  The results of these evaluations showed the effectiveness of our framework, demonstrating that it can provide a context-aware AR experience in a variety of real scenes.
} % end of abstract

\CCScatlist{
  \CCScatTwelve{Human-centered computing}{Human computer interaction (HCI)}{Interaction paradigms}{Mixed / augmented reality};
  \CCScatTwelve{Artificial intelligence}{Computer vision}{Computer vision tasks}{Scene understanding}
}

\begin{document}


\firstsection{Introduction}

\maketitle

\begin{figure*}[t]
  \centering 
  \includegraphics[width=2\columnwidth]{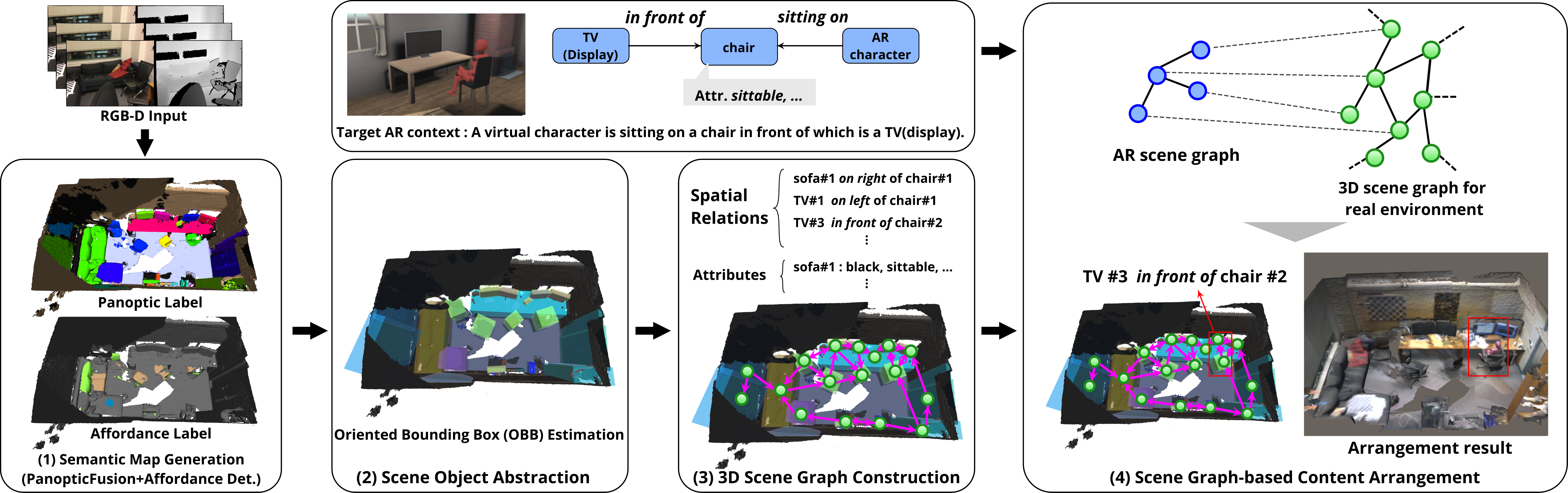}
  \caption{Overview of the proposed framework.}
  \label{fig:system_view}
\end{figure*}

%% \section{Introduction} %for journal use above \firstsection{..} instead
With the recent advances in computer vision and graphics technologies, natural fusion of the virtual and real worlds by Augmented Reality (AR) / Mixed Reality (MR) systems is reaching a practical level.
To integrate virtual objects into real environments accurately and naturally, a deep understanding of physical environments becomes even more crucial for AR/MR systems.
% 近年のコンピュータビジョンやグラフィクス分野における技術的進歩により，仮想世界と現実世界を融合するAR/MRシステムは実用的水準に到達しつつある．
% そして，AR/MRにおいて仮想コンテンツを現実シーンに適切かつ自然に融合するためのシーン理解技術はより重要性を増している．

In conventional approaches, scene understanding in AR has emphasized geometric and photometric registration between the virtual and real worlds\cite{Zhou2008trends, Kim2018revisiting}.
Geometric registration means that virtual content is displayed in a real environment without misalignments.
For example, a virtual entity that is aligned on the ground without floating in the air or that is occluded by real objects gives users a more realistic impression.
Photometric registration makes lighting and shading of virtual objects the same as the real ones, which facilitates the natural representation of virtual objects as well.
Recently, the increased accuracy of camera tracking technology\cite{dai2017bundlefusion, schps2019bad}, such as visual SLAM (Simultaneous Localization and Mapping), and lighting estimation\cite{Bastiaan2017interactive, Rohmer2017natural} has relaxed these registration problems, and it can be said that methods to resolve the difficulties have reached a more practical level\cite{google2017arcore, apple2017arkit}.
% AR分野における現実シーンの理解には，仮想と現実との幾何的整合性及び光学的整合性が伝統的に重要視されてきた\cite{Zhou2008trends, Kim2018revisiting}．
% 幾何的整合性は，仮想コンテンツが現実シーンに位置ずれなく表示されることを意味する．
% 例えば，仮想物体が宙に浮かず現実物体にピッタリ接地していたり，遮蔽されたりすることは，ユーザにより自然な印象を与えてくれる．
% 光学的整合性は，仮想と現実との照明環境や陰影を一致させることを意味しており，幾何学的整合性同様に仮想と現実との自然な表現を促進する．
% これらの整合性問題は，近年のVisual SLAM(Simultaneous Localisation and Mapping)等のカメラトラッキング技術\cite{dai2017bundlefusion, schps2019bad}や照明環境推定技術\cite{Bastiaan2017interactive, Rohmer2017natural}等の高精度化により解消されつつあり，実用水準に達していると考えられる\cite{google2017arcore, apple2017arkit}．

While such relaxation allows ubiquitous AR experiences in a range of settings from specially designed environments to ordinary scenes such as living rooms at home,
 the need for semantic registration to allow more realistic arrangements or interactions of virtual content in different real environments has been raised\cite{Lang2019virtual}.
% Todo: specially designed からリビングまでどこでもできるというニュアンスになっているが，特別な環境からリビングへ持ちだした・広げたというニュアンスの方が適切か？
 Semantic registration means that a defined context for virtual content is preserved and matched to that of a real AR scene as well.
When an AR content developer designs an AR space including a virtual lamp placed on a table, for example, a semantically registered AR space requires placing a virtual light not on a detected plane in the scene, but on the recognized table.
We focus on the understanding of the semantic information and propose a new AR framework that enables virtual content to interact with a real environment while being aware of its context. 
% 上記問題の緩和によって，自宅のリビング等の様々なシーンでARを体験可能になった一方で，近年，様々な現実シーンへの仮想コンテンツの自然な配置やインタラクションを可能にするための仮想と現実との意味的整合性問題が顕在化している\cite{Lang2019virtual}．
% 意味的整合性とは，設定した仮想と現実とのコンテキストを，AR体験する現実シーンにおいても一致させることを意味している．
% 例えばコンテンツ制作者が，テーブル上に仮想ランプが置かれたAR空間をデザインしたのであれば，
% 現実シーン中の検出された平面上ではなく，認識されたテーブル上にそれが置かれることで意味的に整合するAR空間が実現される．
% 我々は，現実シーンの意味的情報を理解することで，コンテキストに即した仮想と現実との自然なインタラクションを実現するARフレームワークを構築を目的とする．

Although object recognition techniques such as semantic segmentation and semantic mapping have been attracting more and more attention by the research community in order to understand real scenes deeply, 
there is little prior work in which AR applications were developed while being aware of rich semantic information.
Commonly used SDKs (ARCore\cite{google2017arcore}, ARKit\cite{apple2017arkit}, and MRTK\cite{ms2016mrtk}) for AR application development provide functions to arrange virtual objects or interact with virtual and real objects using geometry-based scene understanding.
They detect planar regions in a scene from its measured geometric information and assign object categories such as walls, floors, and tables to them.
Such an approach is capable of placing virtual objects onto detected regions. However, it is difficult to deal with interactions in more complex situations that require us to consider the contexts around objects, such as the relations among them, or to discriminate individual objects.
% Semantic segmentationやSemantic mappingなどの物体認識技術が注目を集める一方で，豊かな意味的情報を考慮したARアプリケーションの開発は，事例の少ない萌芽的領域と言える．
% 現在一般的にARアプリの開発に利用されているSDKのARCore\cite{google2017arcore}やARKit\cite{apple2017arkit}，MRTK\cite{ms2016mrtk}では，
% 計測された現実シーンの幾何的情報を基に平面として検出された領域に対して，
% 意味的情報として壁や床，テーブルといった物体カテゴリを当てはめ，仮想コンテンツの配置やインタラクション可能な仕組みを提供している．
% この仕組みでは検出された領域上に仮想コンテンツを配置することは可能だが，周辺物体との関係等のコンテキストを踏まえたより複雑な状況や，
% 個々の物体を区別する必要のある状況での自然なインタラクションの実現は困難である．

On the other hand, some conventional approaches have adopted coordinate sharing between a real space for an AR experience and a virtual space for content creation for context-aware AR\cite{Kato1999artoolkit,Wagner2008pose,Castle2008ptamm,Lepetit2015instant}.
The most fundamental approaches utilize AR markers\cite{Kato1999artoolkit} or image feature tracking\cite{Wagner2008pose}. Such methods set the poses of the virtual object in advance while considering how to present AR content at the time of clue detection.
In this case, AR content designers create AR applications by understanding AR scene contexts based on geometric and semantic information of the scene instead of AR systems.
Also, the recently proposed Visual Positioning System (VPS) \cite{Brachmann2018dsac++, Brahmbhatt2018mapnet} enables content authoring in large-scale scenes. However, these methods have difficulty in dealing with layout change of objects in the scenes, or unknown scenes such as the living rooms of individual users.
% 他方，AR体験する現実シーンの座標系をコンテンツ制作側と体験側とで一致させることで，シーンのコンテキストに応じたARを実現するというアプローチもしばしば採られてきた\cite{Kato1999artoolkit,Wagner2008pose,Castle2008ptamm,Lepetit2015instant}．
% その最も基本的な手法がARマーカの利用\cite{Kato1999artoolkit}や画像特徴の事前記録\cite{Wagner2008pose}によるアプローチであり，それらが認識された際にどのようにコンテンツを提示するのかを仮想オブジェクトのポーズ情報として事前に設定する．
% つまり，ARシステムがシーンの幾何情報や意味情報を基にコンテキストを理解するのではなく，事前にコンテンツ制作者がそれらを理解した上でアプリを作成する．
% また，近年ではVPS(Visual Positioning System)\cite{Brachmann2018dsac++, Brahmbhatt2018mapnet}によって大規模空間でのコンテンツ制作が可能となっている．
% しかし，これらの方法ではシーン中の物体のレイアウトの変化や，ユーザのリビング等の未知シーンへの対応が困難である．

In this paper, we present \textit{Retargetable AR}---a novel AR framework that yields a context-aware AR experience in various scenes based on scene contexts defined by AR content designers and enables natural interaction between the virtual and physical worlds.
The essential feature of the framework is to represent the context of AR content as an abstract 3D scene graph, not as some specific coordinate transformation.
The 3D scene graph symbolizes scene context in the form of a 3D directed graph. The 3D graph consists of nodes corresponding to individual objects and directed edges associated with relations between objects.
Then, by matching the context of AR content with that of the real scene in the 3D scene graphs, we can achieve semantic-aware interactions using virtual content without specific scene coordinates.
In addition, motivated by AR content like a virtual character who takes natural actions in a real environment as humans do (\autoref{fig:teaser_overview}), our framework recognizes additional high-level semantics in real scenes and includes them in the node attributes, which contributes to natural AR representation with interactions.
% 仮想と現実との自然なインタラクションを可能にするARフレームワーク"Retargetable AR"を提案する．
% 重要な鍵となるのは，ARコンテンツの持つコンテキストを具体的な座標変換として表現するのではなく，抽象的な3Dシーングラフとして表現する点である．
% この3Dシーングラフは，個々の物体をノード，それらノード間の関係をエッジとして接続した3D有向グラフによりシーンのコンテキストを表現する．
% そして，ARコンテンツのコンテキストとARを体験する現実シーンのコンテキストとを抽象的な3Dシーングラフ間で一致させることで，特定の座標系に依らず，意味的に整合する仮想コンテンツのインタラクションを実現する．
% 加えて本論文は，\autoref{fig:teaser_overview}のように仮想キャラクタが現実シーンで行動するようなARコンテンツをモチーフとし，現実シーンの高次な意味情報を認識してノードの属性としてグラフに含めることで仮想と現実とが適切かつ自然にインタラクション可能なAR表現を実現する．

To realize such an AR system, it is necessary to recognize not only geometric and semantic information of individual objects but also relations between objects in an online manner.
Our framework rapidly constructs a 3D scene graph by adopting step-by-step scene abstraction.
As shown in \autoref{fig:system_view}, our framework has a four-step structure:
(1) dense 3D reconstruction and semantic information recognition for a scene,
(2) abstraction of scene objects as oriented bounding boxes, 
(3) construction of a 3D scene graph representing the scene context, and
(4) arranging virtual content in a variety of indoor scenes using 3D scene graphs.
% これを実現するためには，現実シーン中の個々の物体の幾何学的・意味的情報のみならず，物体間の関係についてもオンラインで認識する必要がある．
% そこで提案フレームワークでは，段階的にシーンの抽象度を高めることで3Dシーングラフをオンラインで高速に生成する．
% 具体的には，\autoref{fig:system_view}に示すように
% (1)シーンの密な3D復元と意味認識を行い， 
% (2)シーンのコンテキストを表現した3Dシーングラフを構築し，
% (3)3Dシーングラフを用いて様々なシーンへコンテンツを配置する，
% という3ステップの構造を有する．

At the first step of the framework, we generate a 3D semantic map (i.e., a 3D panoptic map with additional affordance information) for the scene geometry and semantic information representation.
We used PanopticFusion\cite{narita2019panopticfusion} as a basis, with modifications for integrating multiple types of image recognition results from RGB-D data into the 3D map (\autoref{fig:system_view}(1)).
The 3D semantic map has voxel-wise resolution and stores object recognition labels and high-level semantic information, like affordances for efficient processing of later steps and a natural interaction representation.
% まず第1ステップでは，RGB-Dデータから多種の画像認識結果を効率的に統合するための改良を加えたPanopticFusion\cite{narita2019panopticfusion}により，シーンの幾何的情報と意味的情報を3次元意味情報マップとして生成する(\autoref{fig:system_view}(1))．
% このマップはボクセルレベルの分解能を持ち，後段の処理の高速化や自然なインタラクション表現のため，物体認識に加えアフォーダンス等の高次の意味的情報を認識する．

Next, we extract a 3D scene graph as a scene context from the 3D semantic map.
To this end, considering computational efficiency, we abstract the individual objects from a voxel representation to oriented bounding boxes (OBBs) and estimate the orientations of OBBs from geometric heuristics and affordances (\autoref{fig:system_view}(2)).
Then, by estimating experimentally defined spatial relations between each object, the system achieves connectivity of the graph (\autoref{fig:system_view}(3)).
% 次に第2ステップでは，この3次元意味情報マップを基にシーンのコンテキストを表現した3Dシーングラフを構築する(\autoref{fig:system_view}(2))．
% 第1ステップで認識された個々の物体をボクセルから軽量な指向性境界ボックスへと抽象化し，更に幾何的なヒューリスティックや高次の意味情報を基にその境界ボックスの向きを高速に推定する．
% そして，経験則に基づき定義した空間的位置関係を推定することでグラフの接続関係を得る．

The final step includes 3D scene graph comparison between the context-represented graph of the AR content and that of the real environment for an AR experience and obtains a context-aware content arrangement from the results (\autoref{fig:system_view}(4)).
Our framework processes the four steps described above in an online fashion according to the RGB-D data input and incrementally updates the 3D semantic map and its corresponding 3D scene graph.
% 最後に第3ステップで，ARコンテンツとして設定したコンテキストと現実シーンのコンテキストをそれぞれ表現したシーングラフを照合することで，
% コンテキストを考慮したコンテンツ配置を実現する(\autoref{fig:system_view}(3))．
% 以上の3ステップを，RGB-Dデータの入力に応じてオンラインで実行することで逐次的に3D意味情報マップと3Dシーングラフを更新する．

The proposed online framework is capable of recognizing individual objects in a real scene, their categories, and high-level semantic information and scene context as spatial relations, which will lead to the realization of rich AR representations in varied environments.
As far as we know, our method achieves AR content arrangements in various real environments based on the deep scene understanding by image-based object recognition and 3D scene graph generation, for the first time.
% 我々のフレームワークは処理全体がオンラインで実行可能であり，個々の物体認識とカテゴリに加え，高次の意味情報及び，シーンのコンテキストを認識することで様々なシーンにおいて高い表現力のARが実現できる．
% 更に本手法は，我々の知る限りにおいて，画像ベースの物体認識に基づく深いシーン理解と3Dシーングラフ生成に基づいて，様々な実環境へのARコンテンツ配置をオンラインで達成する初の手法である．

The contributions of this paper are as follows:
% 本論文のContributionは主に次のとおりである：
\begin{itemize}
  \item We introduce a novel AR framework for context-aware virtual content arrangement in a variety of real environments. 
  Our approach abstracts a scene as a graph representation and places contents in a context-aware way via a graph matching approach, without the need for a pre-defined environment.
  \item We introduce an online 3D scene graph construction method using a 3D semantic map based on image-based object recognition.
  Our method also includes an affordance detection module which makes object orientation estimation easier and allows virtual objects to interact with real environments more naturally.
  \item We experimentally demonstrate that our framework is capable of arranging content in a context-aware way to a variety of interior scenes which have different furniture layouts.
\end{itemize}

% This template is for papers of VGTC-sponsored conferences which are \emph{\textbf{not}} published in a special issue of TVCG.
 
\section{Related Work}
\paragraph{3D Scene Understanding via Semantic Mapping}
Semantic mapping reconstructs the 3D structure of a scene and recognizes its semantics simultaneously, which provides crucial clues for seamless connection between the virtual and physical spaces in an AR experience.
With the great successes achieved by deep convolutional networks in recent years, some works\cite{tateno2017cnn, mccormac2017semanticfusion} have proposed achieving semantic 3D scene reconstruction by combining 2D semantic segmentation and Visual SLAM, and this approach has realized category-level 3D object recognition.
Then, the simultaneous recognition of semantic categories and instances of 3D scene objects has been achieved by using, for example, an object-oriented map representation\cite{runz2018mask, mccormac2018fusion++,xu2019mid}, and a combination\cite{grinvald2019volumetric} of image-based instance segmentation, a famous example of which is Mask R-CNN\cite{he2017mask}, and Visual SLAM.

The most related semantic mapping system, PanopticFusion\cite{narita2019panopticfusion} performed a panoptic segmentation task\cite{kirillov2019panoptic} (i.e., recognizing scene objects at the level of \textit{stuff} and \textit{things} from RGB images) in 3D scenes and reconstructed 3D panoptic maps for the first time.
% Todo: Narita et al proposed ... のような書き方の方が適切かもしれない
This approach enables holistic scene understanding by integrating recognition results into 3D volumes.
This work also demonstrates that geometric information and semantic information at the level of stuff and things of a real environment are useful in AR applications with virtual--real interactions, like a situation in which a virtual character locomotes on the floor and sits on chairs and sofas.
From the perspective of AR, Chen {\it et al.}\cite{chen2018context} have presented a context-aware AR framework with the idea of representing scene context based on dense 3D reconstruction and semantic understanding of scene materials.
% Their framework is able to produce proper collision effects between a virtual object and the real world by recognizing the material of each object.
% 更に，シーンの3次元復元と意味的理解に基づくコンテキストをARに応用した研究として，物体の材質を認識することで仮想と現実との衝突効果を適切に演出するコンテキストに応じたARフレームワーク\cite{chen2018context}が提案されている．
In contrast, we recognize not only the attributes of individual objects but also the relations between them as the scene context and achieve rich AR expressions with a deeper understanding of the holistic scene context. 

\paragraph{AR/VR Experiences in Unknown Environments}
There are previous works to adapt and generate AR/VR experiences in unknown, different environments in real-time\cite{gal2014flare, nuernberger2016snaptoreality, yang2019dreamwalker, hartmann2019realitycheck, cheng2019vroamer}. 
The closest work of our framework is FLARE\cite{gal2014flare}. FLARE generates an optimal layout for AR applications based on the planar geometry of scenes and given constraints. While FLARE focuses on generating simple layouts for AR contents based on geometric information, since our framework utilizes both geometric and semantic information, making it suitable for dealing with a more complex AR scenario/storytelling that is aware of the semantics of individual objects and their relationships.
On the other hand, real-time VR systems like DreamWalker\cite{ yang2019dreamwalker}, VRoamer\cite{cheng2019vroamer} provide continuous, immersive experiences in uncontrolled physical environments. These works are closely related to ours in terms of providing consistency-aware experiences across virtual-real worlds in real-time.

\paragraph{Scene Representation as Graph}
Abstract representations of scene contexts as graph or hierarchical structures based on semantic relations between objects or structures have been utilized in the computer vision and graphics communities due to their compactness and high applicability.
In graphics, abstract representations of scene contexts as relation graph have often been used in the hierarchical analysis of indoor 3D scene models\cite{fisher2011characterizing}, indoor scene synthesis based on various inputs like texts\cite{chang2014learning}, examples\cite{fisher2012example}, and generative models\cite{wang2019planit}.
% シーン全体のコンテキストを物体や構造の意味的な関係に基づくグラフ構造として抽象表現する手法は，そのコンパクトさと応用性の高さからグラフィクスとコンピュータビジョンの両分野で用いられてきた．
% グラフィクス分野においては，屋内3Dシーンモデルを階層的な構造として解析・理解するための研究\cite{fisher2011characterizing, liu2014creating}や，text-based\cite{chang2014learning,ma2018language},example-based\cite{fisher2012example, huang2016structure}，generative model-based\cite{wang2019planit}などの
% 様々な入力を基に新しい屋内3Dシーンモデルを合成する研究において活用されてきた．

In computer vision, the large-scale annotated dataset Visual Genome\cite{krishnavisualgenome} has been introduced in recent years, and has spurred research on scene graph generation tasks\cite{xu2017scene, li2017scene} which enable deep understanding of real scenes from a single RGB image.
Scene graphs are suitable for various vision tasks, such as image retrieval\cite{johnson2015retrieval}, image generation\cite{johnson2018image}, visual question answering\cite{zhang2019empirical}, and image captioning\cite{yao2018exploring}, and scene graph generation is a promising research topic that is attracting more and more attention in these communities. 
The closest works to our approach are recent 3D scene graph generation methods for indoor environments using RGB-D sensing data\cite{armeni20193d, kim2019graph3d,zhou2019scenegraphnet}. 
% コンピュータビジョン分野においては，大規模なアノテーション済みデータセットVisual Genome\cite{krishnavisualgenome}の登場を皮切りとして，一枚の画像からシーンを深く理解することを目的としたシーングラフ生成\cite{xu2017scene, li2017scene, li2018factorizable,yang2018graph, zellers2018neural}に関する
% 研究が近年盛んに行われている．生成されたシーングラフは画像検索\cite{johnson2015retrieval}，画像生成\cite{johnson2018image}，VQA\cite{zhang2019empirical}，Image Captioning\cite{yao2018exploring}といった幅広いvision taskに応用されており，
% 非常に注目度が高い発展が約束された研究領域である．
% 最近は，屋内実環境においてRGB-D画像を用いて3次元空間に対する3Dシーングラフを生成する手法\cite{armeni20193d, kim2019graph3d,zhou2019scenegraphnet}も提案されてきている．

In our framework, in terms of scalability and flexibility, we represent a scene context as a 3D scene graph, similarly to these methods. 
The graph is constructed in an online manner based on a 3D semantic map reconstructed from incoming RGB-D images of the real environment.
To the best of our knowledge, this is the first AR framework based on a 3D scene graph generated online.
% 我々のフレームワークでは，拡張性と柔軟性の観点からこれらの研究と同様にシーンのコンテキストをシーングラフとして表現し，
% RGB-D画像から実環境を計測して得られる3D意味的情報マップを基にオンラインで3Dシーングラフを構築する．
% これにより，我々の知る限りにおいて世界初のオンライン生成された3Dシーングラフに基づくARフレームワークを実現する．

\section{Proposed Framework}
The proposed framework takes in streams of RGB-D images from off-the-shelf scanners and generates a 3D semantic map from the images (\autoref{fig:system_view}). With a 3D scene graph constructed from the map and a scene graph characterizing the context of AR content defined by content designers, the framework outputs context-aware AR content arrangements according to a real environment.
% With a 3D scene graph constructed from the map and a scene graph characterizing the context of AR content defined by content designers, the framework outputs context-aware AR content arrangements according to a real environment.
% The semantic map generation module employs PanopticFusion as a basis and is equipped with a multi-label integration extension that enables efficient integration of inferred additional semantic information along with object recognition results into voxel spaces.
Diverse types of semantics (e.g., affordance, material, texture) could be considered as additional semantic information here. In this paper, we select affordance, which is closely related to interactions in real environments, as an additional semantic.
% The 3D semantic map is incrementally updated with incoming RGB-D images, and the 3D scene graph construction module at the next step extracts a scene graph as the scene context representation.
% The 3D scene graph construction module estimates abstracted shapes, positions, and poses of scene objects as OBBs. The orientations of the OBBs are efficiently determined using geometric heuristics or affordance information.
% Then the module extracts spatial relations between objects and constructs a 3D scene graph that has a set of nodes corresponding to instances and edges between nodes associating relations with nodes.
% Finally, the content arrangement module compares the constructed graph with an AR scene graph defined by content designers, obtaining AR content arrangements whose contexts are semantically consistent, as the designers intended.
% We describe the details of each module in our framework below.
Each module will be described in details in the following subsections.
% 我々のフレームワークはRGB-D画像系列を入力とし，ARコンテンツ制作者が定義するコンテキストを表現したシーングラフを基に，現実シーンへそのコンテキストに応じたコンテンツ配置を実現する(\autoref{fig:system_view})．
% 意味的情報マップ生成モジュールには，PanopticFusionを基礎として物体認識に加えて画像ベースの認識器で推論された高度な意味的情報を効率的にボクセル空間へ統合可能な仕組みを導入した．
% また高度な意味情報として，仮想と現実とのインタラクションに関連性の高いアフォーダンスを認識することで，両者の自然な融合を実現する．
% %ボクセル空間に統合された個々の物体単位での幾何的・意味的認識情報から，容易に物体毎に意味的情報と結びついたメッシュを抽出することができる．
% 意味的情報マップは入力画像情報に応じて逐次的に更新され，次段の3Dシーングラフ生成モジュールにより，シーンのコンテキストを表現した3Dシーングラフが構築される．
% ここで，物体の情報をボクセルから指向性境界ボックスとして抽象化し，各境界ボックスの正面方向を物体の形状やアフォーダンスを考慮して高速に推定する．
% そして，抽象化された境界ボックスの位置と姿勢を基に，各物体間の空間的な位置関係を推定することで，
% 各物体をノード，ノード間の関係をエッジとして接続した3Dシーングラフを構築する．
% 最後にコンテンツ配置モジュールにおいて，上記コンテンツ制作者が定義したARシーングラフと意味的情報マップから構築した3Dシーングラフとを照合をすることで，
% 制作者が意図したコンテキストと現実のコンテキストが意味的に整合したコンテンツ配置を実現する．

% 以下では，提案フレームワークの各モジュールについて詳細を述べる．

\subsection{Semantic Map Generation}
\subsubsection{Multi-label Integration Module}
\label{sec:multi-label}

\begin{figure}[t]
  \centering 
  \includegraphics[width=\columnwidth]{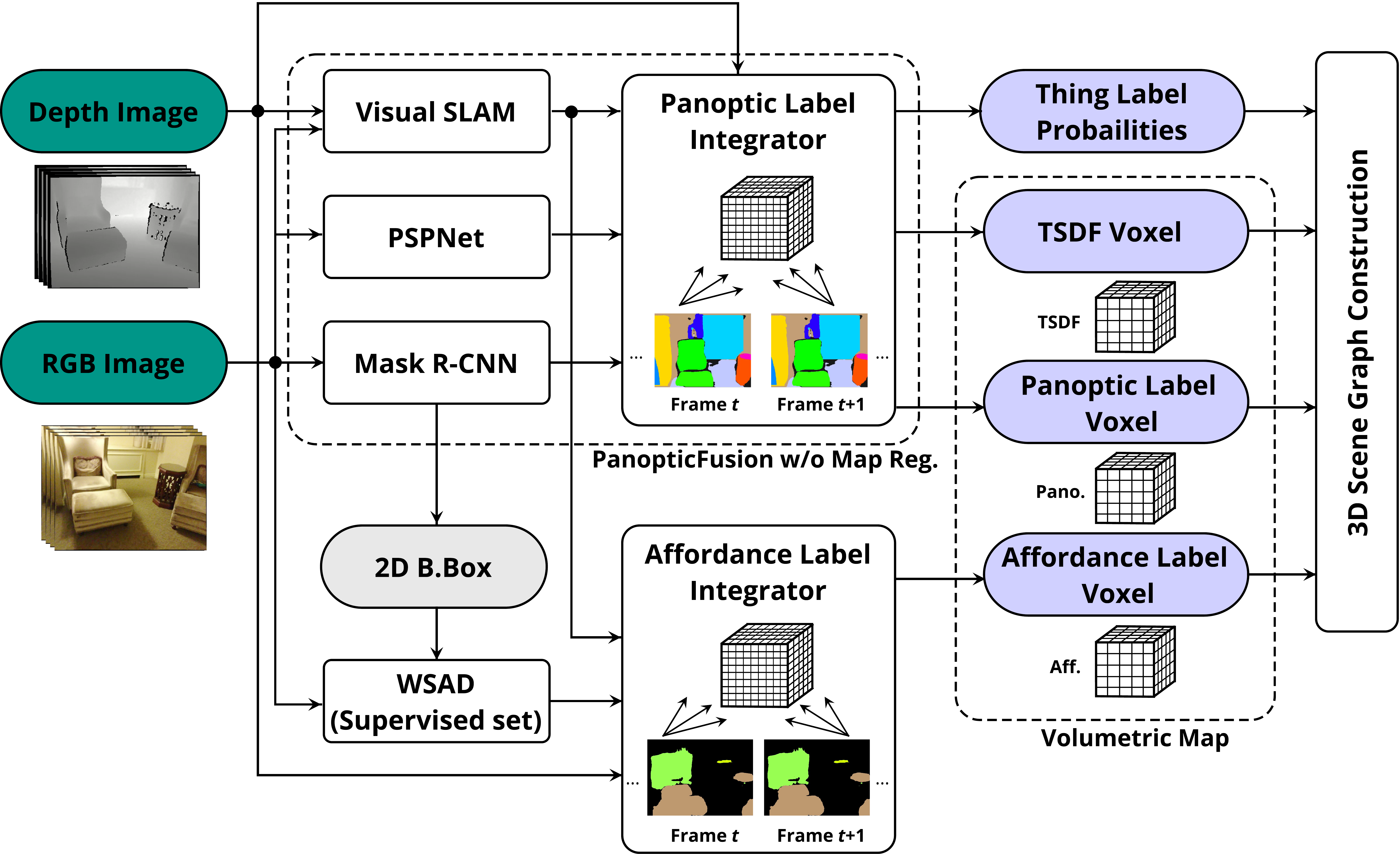}
  \caption{Overview of semantic map generation module.
  %  The volumetric map preserves TSDF and other semantic labels separately so that it can efficiently accommodate additional classification modules.
  % The affordance detection module uses 2D bounding boxes of Mask R-CNN to apply per object affordance segmentation.
  }
  \label{fig:panoptic_aff}
\end{figure}

As shown in \autoref{fig:panoptic_aff}, we map a Truncated Signed Distance Function (TSDF) representing the scene geometry and object recognition results into 3D voxel spaces based on camera poses estimated from RGB-D data and depth values, the same as PanopticFusion\cite{narita2019panopticfusion}. TSDF enables implicit volumetric representation of the scene by storing the truncated signed distance to the object surface\cite{curless1996volumetric}, and allows easily extracting meshes of objects, which are used for the next object abstraction step, using Marching Cube algorithm\cite{lorensen1987marching}.

In the PanopticFusion pipeline, each voxel contains truncated signed distances to the object surface, and panoptic labels, representing stuff labels and things labels uniformly, are integrated into it together with truncated signed weights.
By contrast, to perform efficient addition and integration of higher-level semantic information, we adopt a structure for preserving TSDF, panoptic labels, and additional semantic labels from other recognition results in independent voxel spaces.
This data structure allows for efficient multi-label integration. While TSDF voxels are updated in the whole scene, including spaces that do not contain objects, through raycasting, it is sufficient for the voxels storing semantic labels to be updated only around the object surface,
which reduces the update processing cost and the number of voxels for hashing and enables us to integrate geometry and semantics separately.
Besides, this approach easily provides 3D semantic meshes by obtaining the correspondences of voxel spaces preserving each semantic label and mesh generated from the TSDF.
To simplify the implementation, we omit the map regularization introduced in the original method, which is based on a Conditional Random Field (CRF).
% \autoref{fig:panoptic_aff}に示すように，我々はPanopticFusion\cite{narita2019panopticfusion}と同様に，
% RGB-D画像から推定したカメラポーズと奥行き値を基に，形状を表すTSDF(Truncated Signed Distance Function)と物体認識結果を3Dボクセル空間にマッピングする．
% PanopticFusionでは物体表面までの打ち切り符号付き距離を格納したボクセルに，
% stuffとthingsを統一的に表したpanoptic labelが打ち切り符号付き重みと共に統合される．
% これに対して，我々は高次の意味的情報を効率的に追加・統合可能にするため，
% TSDFとpanoptic label及び，その他の認識器からの意味ラベルをそれぞれ独立のボクセル空間に保持する構造とした．
% これにより，TSDFボクセルを光線追跡で物体が存在しない空間についても更新する一方で，
% 意味ラベルを保持するボクセルについては物体表面周辺のみラベルとその重みを更新すれば十分であるため，
% 更新処理コストの削減とハッシュに追加するボクセル数の低減が可能となり，幾何的情報とは独立に意味的情報を効率的に統合できる．
% また，TSDFから生成したメッシュを各ラベルが格納されたボクセル空間に対応付けることで容易に意味情報付きメッシュが得られる．
% なお本研究では，PanopticFusionのConditional Random Fieldによるマップ正則化処理は実装の簡略化のため省略した．

\subsubsection{Affordance Detection Module}
\label{sec:affordance}

\begin{figure}[t]
  \centering 
  \includegraphics[width=\columnwidth]{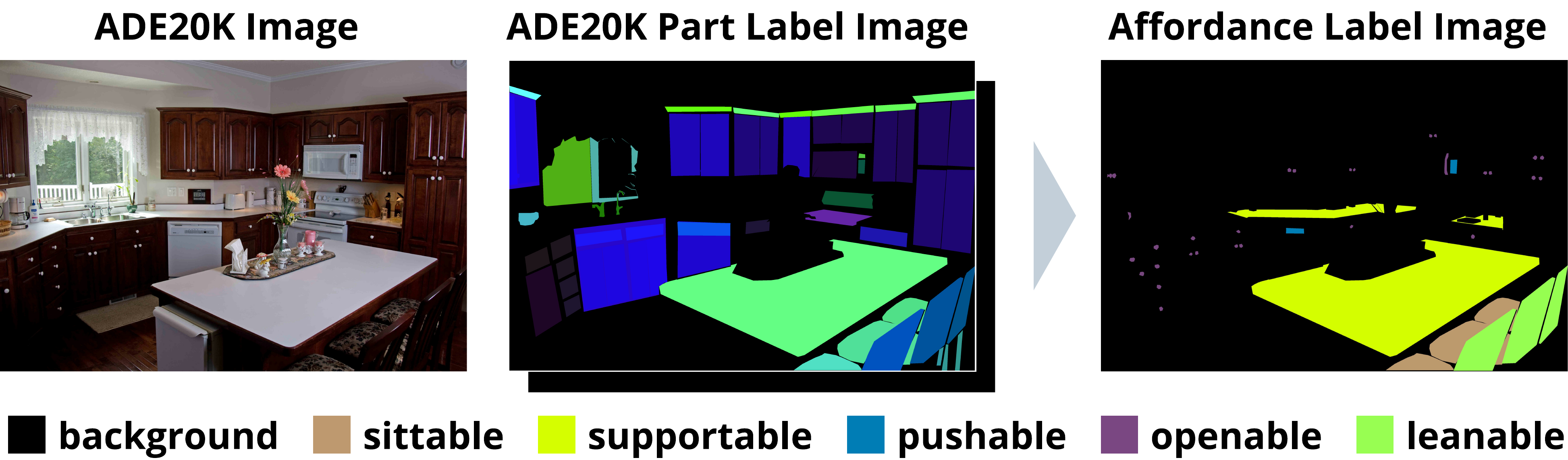}
  \caption{Indoor scene affordance examples converted from ADE20K parts labels. }
  \label{fig:affordance_example}
 \end{figure}

Recognizing affordances (i.e. interactions that real objects offer virtual contents) will contribute to natural interactions.
In recent years, some learning-based methods that detect object affordances from RGB images via neural networks\cite{luddecke2017learning, sawatzky2017weakly, do2018affordancenet} have been proposed.
However, most of the open datasets with annotated affordance labels are mainly built for supporting robot manipulation in different scenes and are not matched with our purpose, like assisting interactions with virtual content in indoor scenes such as the living rooms of users.
Therefore, inspired by the method\cite{luddecke2017learning} of Luddecke {\it et al.}, we built an ADE20K Affordance dataset by converting the part labels of ADE20K\cite{zhou2017scene} into affordance labels, assuming the indoor scene interactions shown in \autoref{fig:affordance_example}.
We use the fully supervised setting of Weakly Supervised Affordance Detection (WSAD)\cite{sawatzky2017weakly} as an affordance detection head trained on the created dataset.
The affordance detector is applied to RGB images in the input RGB-D streams and outputs pixel-wise affordance segmentation of these images.
Note that the segmentation process is applied not on the whole image but the cropped regions of interest using the bounding boxes estimated for individual objects by Mask R-CNN.
% 実物体が仮想コンテンツに対して許容するインタラクションを意味するアフォーダンスを認識することは自然なインタラクションの実現へ繋がる．
% 近年，RGB画像からneural networkを用いて物体のアフォーダンスを学習ベースで認識する手法が提案されている\cite{luddecke2017learning, sawatzky2017weakly, do2018affordancenet}．
% しかし，アフォーダンスラベルが付与された公開データセットの多くはロボットのマニピュレーション支援を目的としており，
% 我々がモチーフとするような，リビング等の屋内における仮想コンテンツの様々なインタラクションを補助する目的に適さない．
% このため，Luddeckeら\cite{luddecke2017learning}の手法に触発され，ADE20K\cite{zhou2017scene}データセットの部位ラベルを，\autoref{fig:affordance_example}のように屋内シーンにおけるインタラクションを想定したアフォーダンスラベルへと変換したデータセットを構築した．
% 作成したデータセットを用いてWeakly Supervised Affordance Detection(WSAD)\cite{sawatzky2017weakly}を教師あり学習させて，入力されるRGB-D画像のうちのRGB画像に対して画素単位でのアフォーダンスセグメンテーションを行う．
% ただし，アフォーダンスセグメンテーションは画像全体ではなく，Mask R-CNNで検出された個々の物体の境界ボックスを用いて切り抜いた物体領域毎に適用する．

% In this paper, we add the affordance detection module to the modified PanopticFusion to realize the multi-label integration described in the previous section.
% Following the pipeline in \autoref{fig:panoptic_aff}, panoptic and affordance segmentation results for RGB sequences are incrementally integrated into a voxel-based volumetric semantic map.
% 前節で述べたmodified PanopticFusionに，アフォーダンス認識モジュールを追加することで
% \autoref{fig:panoptic_aff}に示すように入力RGB画像に対するpanoptic segmentationとaffordance segmentationの認識結果がボクセル単位の3D意味的情報マップとして統合される．

\subsection{Scene Object Abstraction}
\label{sec:scene_object_abstraction}
In general, dense reconstruction and labeling of scenes are desirable for virtual content locomotion or interaction in the scenes.
However, with respect to comparing the real environments with the context assumed by the virtual content designers, it is not suitable to deal with the scene representation as dense voxels due to the processing complexity and computational cost involved.
Also, to ensure the online performance of the framework, it is necessary to represent the scene context information in a more tractable way.
% 仮想コンテンツがシーンを移動したり実物体に対してインタラクションしたりするうえでは，シーンの幾何的及び意味的情報は密に取得できていることが望ましい．
% しかし，現実シーンとARコンテンツ制作者の想定したコンテキストとの比較を行う観点では，
% シーンを密なボクセル表現のまま取り扱うことは，処理の複雑化や計算コストの増大を招き，オンライン動作を確保する面でも不都合である．

\begin{figure}
  \centering 
  \includegraphics[width=\columnwidth]{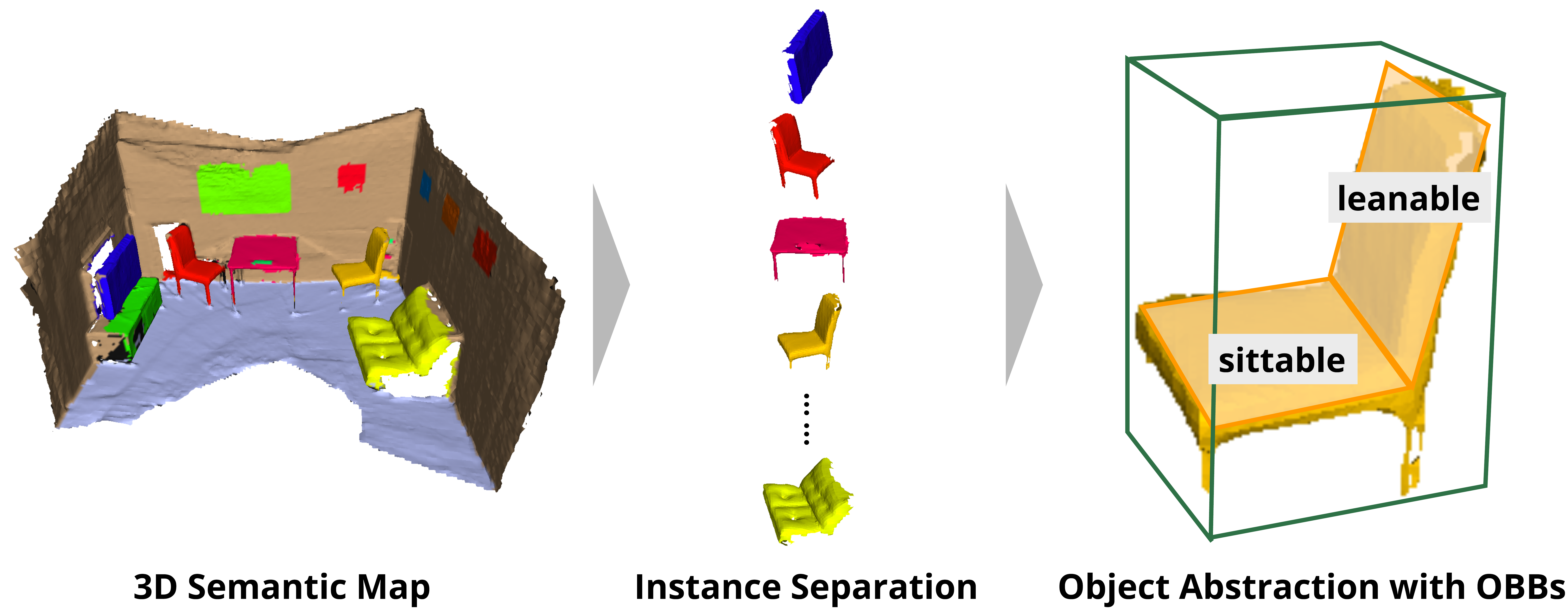}
  \caption{Abstracting scene objects with Oriented Bounding Boxes (OBBs) per instance. Affordance parts of objects will be abstracted as well, if recognized.}
  \label{fig:instance_sepration}
 \end{figure}

Here, taking the 3D scene graph construction process at the later step into account, we abstract individual object poses and positions as a lightweight 3D representation on the basis of the obtained map.  
In particular, separating the 3D structures per object, we estimate instances and parts level Oriented Bounding Boxes (OBBs).
Since the semantic map has recognition labels for individual objects, as \autoref{fig:instance_sepration} shows, we can easily separate the 3D structure of each recognized instance from the map.
Then we estimate OBBs for separated instances to abstract their positions, orientations, and sizes. The positions and sizes of OBBs are determined as minimum bounding boxes using center points and points extracted from meshes of separated instances. The orientations are estimated using geometric heuristics described later.
The OBBs for the regions with affordance are also estimated and stored as parts separately with the instance ones.
Our framework assumes that most of the objects in general indoor environments are standing upright, so that OBB instances are estimated as vertical axis-aligned (z-axis aligned) ones. An OBB corresponding to a part of an object does not employ this assumption.
%  そこで，後述する3Dシーングラフ構築処理を見据えて，取得した意味情報マップを基に個々の物体の位置・姿勢を軽量で取り扱いやすい3次元表現として抽象化する．
% 具体的には，各物体の3D構造を分離して物体及び部位単位での指向性境界ボックス(OBB)を推定する．
% 意味情報マップには個々の物体の認識ラベルが付されているため，\autoref{fig:instance_sepration}のように容易に意味情報マップを分割して識別された物体毎の3次元構造に分離することができる．
% 我々は分離された各物体に対して指向性境界ボックスを推定し，抽象化された物体の位置姿勢，寸法を表現するパラメータを得る．
% 同様にして，高次認識ラベル（本論文ではアフォーダンス認識ラベルに相当する）が得られた領域の3次元構造も抽出できるため，
% 物体全体に対する境界ボックスとは別にその部位を表現する境界ボックスを推定する．
% なお，本論文では，一般的な屋内シーンを想定した際に物体の多くは正立していることを考慮して各インスタンスの指向性境界ボックスは鉛直軸(z軸)が揃っていると仮定する．
% 物体に含まれる部位に対応する境界ボックスについては，その仮定に依らず指向性境界ボックスを計算する．

We use several geometric heuristics to predict the orientations of the OBBs.
First, we select the front and back axis of an OBB from its extension for each category. 
For instance, most of the furniture objects in an indoor scene, which we mainly focus on, are longer in the left/right direction than the front/back direction.
In such a situation, it is appropriate to set an axis corresponding to the front and back to be perpendicular to the direction of the longer extension. In some categories, such as beds, we set it otherwise.
Next, we utilize the observation that most of the recognized objects have more measured surfaces on the front side than the back. 
We determine the orientation of an OBB as the normal orientation of its surface that is closest to the major direction of normals for recognized objects.
% 我々は指向性境界ボックスの正面方向をいくつかの幾何的なヒューリスティックスを基に推定する．
% 今回我々が対象とする物体は主に屋内に存在する家具であり，その多くは横長の形状を持っているということを考慮し，
% 境界ボックスの短軸方向が前後方向に対応すると仮定する．
% 更に，認識された各物体は正面側の計測点の方が多いという仮定を置き，計測点の重心が境界ボックスの中心より正面側に存在することを利用して，正面側へと伸びるベクトルと同じ側を向くように境界ボックスの正面方向を決定する．

For some categories, our framework is able to utilize affordance information to estimate the front side of an OBB more accurately.
Specifically, the orientations of some objects that have \textit{sittable} and \textit{leanable} areas, such as chairs and sofas, are robustly and quickly estimated from the positions of the OBBs of those affordance labels.
The OBB for \textit{sittable} is usually located in front of the \textit{leanable} one, and the direction to the \textit{sittable} area, i.e. the seat area, from the \textit{leanable} area, i.e. the backrest area, is always the same as the front orientation.
We calculate the direction vector corresponding to the front orientation using the positions of \textit{sittable} and \textit{leanable} OBBs and set the orientation of an instance OBB as the normal orientation of its surface that is closest to the direction vector.
% 上記のヒューリスティックスにより妥当な向きが求められない縦長物体の多いカテゴリ(ベッド etc.)や特定のアフォーダンスsittable, leanableを持つカテゴリ(椅子，ソファ etc.)の物体については正面方向をより正しく推定するためのヒューリスティックスを用いて境界ボックスの向きをrefineした．
% 縦長の物体については上述したヒューリスティックスにおいて長軸方向を前後方向に対応させればよい．
% sittableとleanable領域を持つ物体についてはその正面方向を高速かつ頑健に推定できる．
% 具体的には，sittable領域が必ずleanable領域の前に存在することを利用し，両者の境界ボックスの位置関係から正面方向に対応する方向ベクトルを求め，その方向に対応する面の方向を境界ボックスの正面方向とする．

\subsection{3D Scene Graph Construction}
% \begin{figure}[t]
%   \centering 
%   \includegraphics[width=\columnwidth]{scene_graph_examples.pdf}
%   \caption{A visualization example of a 3D scene graph from a 3D semantic map, showing some of the relations for simple visualization (left:\textit{near}, right:\textit{in front of}).}
%   \label{fig:scenegraph_real}
% \end{figure}

\begin{figure}
  \centering 
  \includegraphics[width=\columnwidth]{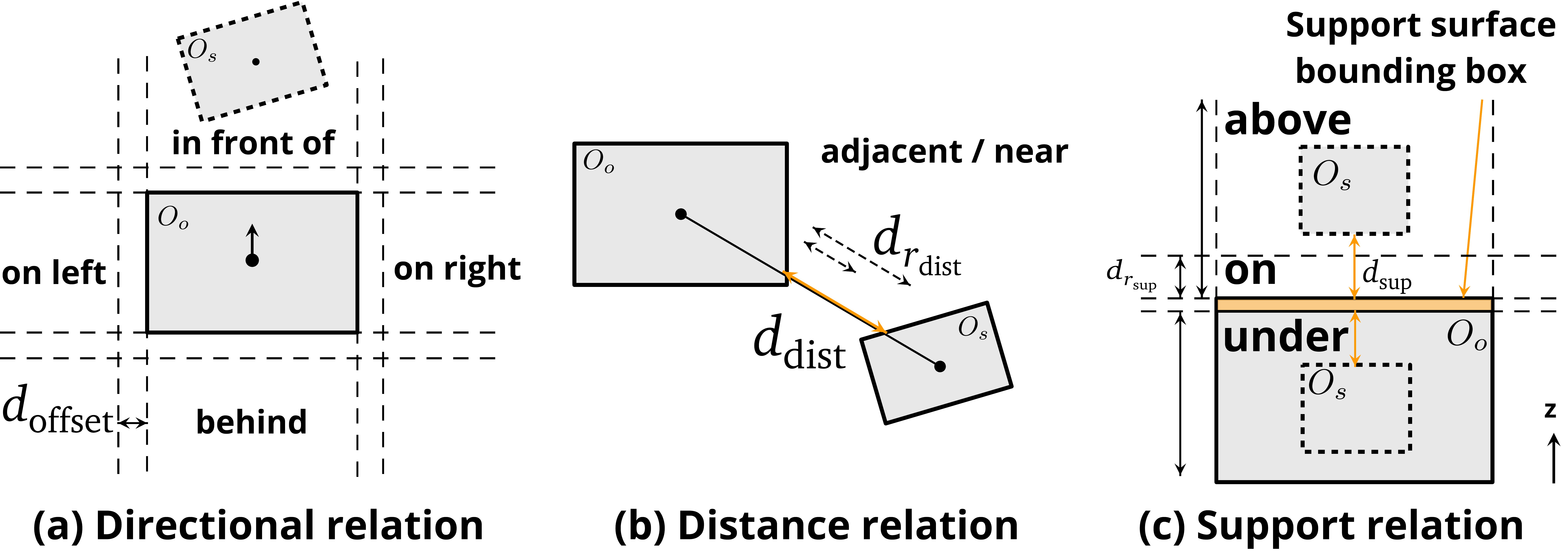}
  \caption{Spatial relations. The relations are defined with a pair of object bounding boxes.}
  \label{fig:spatial_rel_overview}
\end{figure}

\subsubsection{Graph Definition}
\label{sec:graph_definition}
To represent scene context, we introduce a 3D directed graph $G = (V,E)$ that encodes recognized objects as nodes $v_i \in V$, and one or more relations $r(v_i, v_j) \in R$ between nodes as edges $e_{ij} \in E$. 
The node $v_i$ corresponds to an instance OBB described in \autoref{sec:scene_object_abstraction} and holds diverse information concerning each object as attributes $a_{v_i} \in A$. The attributes could include diverse information, such as color, size, and material, in addition to affordance, which our framework employs, depending on the assumed context for the AR content.
% \autoref{fig:scenegraph_real} shows a visualization example of a 3D scene graph constructed online from a 3D panoptic map.
% 我々は現実シーンのコンテキストの表現のため，シーン中に識別された個々の物体をノード$v_i \in V$，1つ以上の関係$r(v_i, v_j) \in R$を持つ物体ノード$v_i, v_j$間のエッジを$e_{ij} \in E$として符号化した3次元有向グラフ表現$G = (V,E)$を導入する．
% 物体ノード$v_i$は\ref{sec:scene_object_abstraction}項で抽象化された各物体に対応しており，属性$a_{v_i} \in A$としてその物体に関する様々な情報を持つ．本論文で我々が利用するアフォーダンス領域の情報の他にも，例えば，色，大きさ，材質等の情報をコンテンツ制作者の想定するコンテキストに応じて保持しうる．
% % 物体$v_i$の特定の部位$k$（例えば，椅子における背もたれ）が識別されて境界ボックスを持つ場合は，物体と同様に部位についてもノード$v_k$で表現し，$v_i$と$v_k$は後述する部位関係を持つエッジ$e_{ik}$で接続される．
% \autoref{fig:scenegraph_real}はオンラインで生成された3D panoptic mapから得られる3Dシーングラフを一部の関係について可視化したものである．

\subsubsection{Spatial Relations}
The proposed framework utilizes the graph representation defined in \autoref{sec:graph_definition} and extracts relations between nodes $r(v_i, v_j) \in R$ as spatial relations between abstracted objects in real environments.
% In contrast to the scene graph generation from 2D image, the relations in a 3D space should have view-independent consistency. 
We adopt a geometric rule-based approach to extract spatial relations using the compactly parameterized shape and pose of OBBs. These relations allow to efficiently construct a 3D scene graph.
In particular, we extract spatial relations $r(v_s, v_o)$ between the subject $v_s$ and the object $v_o$ as directed-relations $r(O_s, O_o)$ derived from their OBBs $O_s, O_o$. 
% 提案フレームワークでは，\ref{sec:graph_definition}項で定義したグラフ表現を利用して，各ノード間の関係$r(v_i, v_j) \in R$を\ref{sec:scene_object_abstraction}項で抽象化したシーン中の物体間の空間的な位置関係として抽出する．
% 2次元画像のシーングラフでの空間的な位置関係とは異なり，3次元空間に対して与える関係は特定の視点の依存しない一貫性を持つ必要がある．そこで，各物体の指向性境界ボックスとしてコンパクトにパラメータ表現された形状や姿勢から，単純な幾何的なルールを用いて関係を抽出し，3次元空間上で抽出した関係を接続することで効率的に3次元シーングラフを構築する．
% 具体的には，親物体$v_s$と子物体$v_o$の間に定義される空間的な関係$r(v_s, v_o)$を，それぞれの持つ境界ボックス$O_s, O_o$から定まる有向関係$r(O_s, O_o)$として抽出する．

% A number of spatial relations in 3D space have been explored, from simple relations for object pairs to group relations for multiple objects\cite{wang2019planit,li2017generating, ma2018language}.
Taking the online performance into account, we define the following fundamental spatial relations $r (O_s, O_o)$ which just need relatively lightweight processing based on the ones proposed by Wang {\it et al.} \cite{wang2019planit} and Li {\it et al.}\cite{li2017generating}.
% 3次元空間における空間的な位置関係は物体組に対する単純なものから，複数の物体に対する関係まで様々なものが探索されてきた[]．
% 我々は現実シーンのグラフをオンラインで構築することを念頭におき，指向性境界ボックスから比較的軽量なヒューリスティック処理により抽出できる基本的な関係$r (O_s, O_o)$をWangら\cite{wang2019planit}やLiら\cite{li2017generating}の提案する関係を基に次のように定義する：

\paragraph{\it{Directional Relation} $(r_\mathrm{dir})$.}
Denotes the direction in which $O_s$ is located relative to $O_o$.
As shown in \autoref{fig:spatial_rel_overview}(a), 
when the center point of $O_s$ is placed in regions between lines defined by extensions of $O_o$ in the horizontal plane (XY-plane), we represent relations according to the orientation of $O_o$ as 
$\left\{r_\mathrm{dir}(O_s, O_o) |r_\mathrm{dir} = \it{in \ front \ of, behind, on \ left, on \ right} \right\}$.
$d_\mathrm{offset}$ here means an offset parameter for extending $O_o$ and enlarging valid areas.
% \paragraph{\it{Directional Relation} $(r_\mathrm{dir})$.}
% $O_s$が$O_o$に対して位置する方向を示す関係である．
% \autoref{fig:spatial_rel_overview}(a)のように
% 水平面における$O_o$の辺の延長線同士に挟まれる領域内に$O_s$の中心点が含まれる場合に$O_o$の向きに応じて$\left\{r_\mathrm{dir}(O_s, O_o) |r_\mathrm{dir} = \it{in \ front \ of, behind, on \ left, on \ right} \right\}$と表す．
% $d_{\it{offset}}$は$O_o$を拡大して探索範囲を広げるためのパラメータとする．

\paragraph{\it{Distance Relation} $(r_\mathrm{dist})$.}
Denotes the distance between $O_s$ and $O_o$.
As shown in \autoref{fig:spatial_rel_overview}(b),
when the distance between the surfaces of $O_s$ and $O_o$ on the line connecting the centers of $O_s$ and $O_o$ in the horizontal plane (XY-plane) is shorter than a defined threshold $d_{r_\mathrm{dist}}$,
we represent the relations as $\left\{r_\mathrm{dist}(O_s, O_o) |r_\mathrm{dist} = \it{near, adjacent} \right\}$.
Note that distance relations are bidirectional; that is, $r_\mathrm{dist}(O_s, O_o)$ and $r_\mathrm{dist}(O_o, O_s)$ always works at the same time.
% \paragraph{\it{Distance Relation} $(r_\mathrm{dist})$.}
% $O_s$と$O_o$との距離を示す関係である．
% \autoref{fig:spatial_rel_overview}(b)のように
% 水平面における$O_s$と$O_o$の中心を結ぶ直線上で，
% $O_s$と$O_o$の表面間の距離が予め定めたしきい値$d_{r_\mathrm{dist}}$より小さい時に$\left\{r_\mathrm{dist}(O_s, O_o) |r_\mathrm{dist} = \it{near, adjacent} \right\}$と表す．
% なお，$r_\mathrm{dist}(O_s, O_o)$が成立する時，$r_\mathrm{dist}(O_o, O_s)$も成立する双方向的な関係である．

\paragraph{\it{Support Relation} $(r_\mathrm{sup})$.}
Denotes spatial relations on the vertical axis (z-axis) between $O_s$ and $O_o$. 
As shown in \autoref{fig:spatial_rel_overview}(c),
the relations are defined by the distance between the support surface bounding box of $O_o$ and top or bottom planes of $O_s$ with threshold $d_{r_\mathrm{sup}}$ and represented as
$\left\{r_\mathrm{sup}(O_s, O_o) |r_\mathrm{sup} = \it{on, above, under} \right\}$.
Here, the center of $O_s$ is included in $O_o$ in the horizontal plane.
The support surface bounding box is specified per $O_o$ to extract the support relations of each object pair.
When the object corresponding to $O_o$ has \textit{sittable} or \textit{supportable} areas, we use their OBBs as the support surface bounding boxes; otherwise we utilize $O_o$ itself.
% \paragraph{\it{Support Relation} $(r_\mathrm{sup})$.}
% $O_s$と$O_o$の鉛直(z軸)方向の位置関係を示す関係である．
% \autoref{fig:spatial_rel_overview}(c)のように
% $O_o$のsupport surfaceと$O_s$の上面或いは底面との距離に対する
% しきい値処理$d_{r_\mathrm{sup}}$により決定し，$\left\{r_\mathrm{sup}(O_s, O_o) |r_\mathrm{sup} = \it{on, above, under} \right\}$と表す．
% ただし，$O_s$の中心点は$O_o$にxy平面上で含まれるものとする．
% support surfaceは各物体について鉛直方向の関係を推定するために定義された指向性境界ボックスであり，sittable或いはsupportableとして認識された領域がある場合はその指向性境界ボックスを，それ以外の場合は$O_o$を利用する．

% \paragraph{\it{Part Relation} $(r_{part})$.}
% 物体$O_s$の部位として検出された領域を$O_o$とした時に$O_s$と$O_o$を接続する関係である．
% 本論文におけるアフォーダンス領域のように物体の一部分が3次元位置構造と共に認識されて境界ボックスを持つ\autoref{fig:spatial_rel_overview}(d)のような場合に$\left\{r_{part}(O_s, O_o) |r_{part} = \it{has} \right\}$として表す．

\

All relations $r(O_s, O_o) \in R$ defined above are investigated for each pair of OBBs $(O_i, O_j)$.
If none of the relations are detected, the object pair does not have edges between them in the 3D scene graph.
% 上記のように定義した全関係$r(O_s, O_o) \in R$を個々の物体の指向性境界ボックスの全組み合わせ$(O_i, O_j)$に対して判定する．
% いずれの関係にも該当しない物体組はその間にエッジを持たない．

\subsection{Scene Graph-based Content Arrangement}

\paragraph{\it{Content Description.}}
As described above, our framework represents the context of AR content as a scene graph, called an AR scene graph.
Therefore, to achieve natural behaviors of AR contents in real scenes in which they are semantically registered to their AR scenarios,
it is required for content designers to define in advance what situations of scenes, or contexts, are needed for the scenarios.
% 我々のフレームワークでは，これまでに述べた通りARコンテンツのコンテキストをシーングラフとして表現する．
% 従って，ARコンテンツの想定するシナリオに対して意味的に整合性を取れた自然な振る舞いをさせるためには，
% どういったシーンの状況，コンテキストが必要かということをコンテンツ制作者が事前に定義する必要がある．

AR scene graphs are also configured as $G' = (V', E')$ following the format defined in \autoref{sec:graph_definition}. In $G'$, a virtual content is also represented as a node $v'_{c} \in V'$ and is connected with other nodes via relations $r_\mathrm{act} \in R$ denoting their interactions. 
Suppose the AR context is defined as "A virtual character is sitting on a chair in front of which is a TV", as shown in \autoref{fig:teaser_overview}.
First of all, nodes for required components, "TV", "chair" and "virtual character" are added as $v'_0, v'_1, v'_2 \in V'$.
Then we associate TV $v'_0$ and chair $v'_1$ with $\textit{in front of}(v'_0, v'_1)$ as "A TV is in front of a chair" and connect an edge $e'_{01} \in E'$.
Finally, the virtual character $v'_{c}$ and chair $v'_1$ are associated with $r_\mathrm{act}=sitting \ on(v'_{c}, v'_1)$ and $e'_{c1}$ is created in $G'$ for representing the situation that "virtual character" is "sitting on" a "chair".
Our framework compares an AR scene graph $G'$ defined in this manner with a real environment graph $G$ constructed in \autoref{sec:graph_definition} and outputs content arrangements regarding the results.
% ARシーングラフも\ref{sec:graph_definition}項で定義したフォーマットに則って$G' = (V', E')$として設定する．ただし，G'上では仮想コンテンツもノード$v'_{c} \in V'$として表現し，インタラクション行動を表す関係$r_\mathrm{act} \in R$を定義して他のノードと接続する．
% 例えば，事前に設定したコンテキストを\autoref{fig:teaser_overview}のように「仮想キャラクタがテレビを前にした椅子に座っている」とする場合，
% まず，シーンに必須の要素である「テレビ」「椅子」「仮想キャラクタ」を物体ノード$v'_0, v'_1, v'_2 \in V'$として追加する．
% 次に「テレビ」は「椅子」の「正面にある」として，テレビ$v'_0$と椅子$v'_1$を$\textit{in front of}(v'_0, v'_1)$という関係に対応するエッジ$e'_{01} \in E'$を接続する．
% そして，$G'$において「仮想キャラクタ」が「椅子」に「座る」という状態を表すために，仮想キャラクタ$v'_{c}$と椅子$v'_1$を関係$r_\mathrm{act}=sitting(v'_{c}, v'_1)$を持つエッジ$e'_{c1}$を接続する．
% 本フレームワークではこのようにしてARシーングラフ$G'$と現実シーンから取得した3Dシーングラフ$G$とを照合することでコンテンツの配置を行う．

\begin{figure*}[t]
  \centering 
  \includegraphics[width=2\columnwidth]{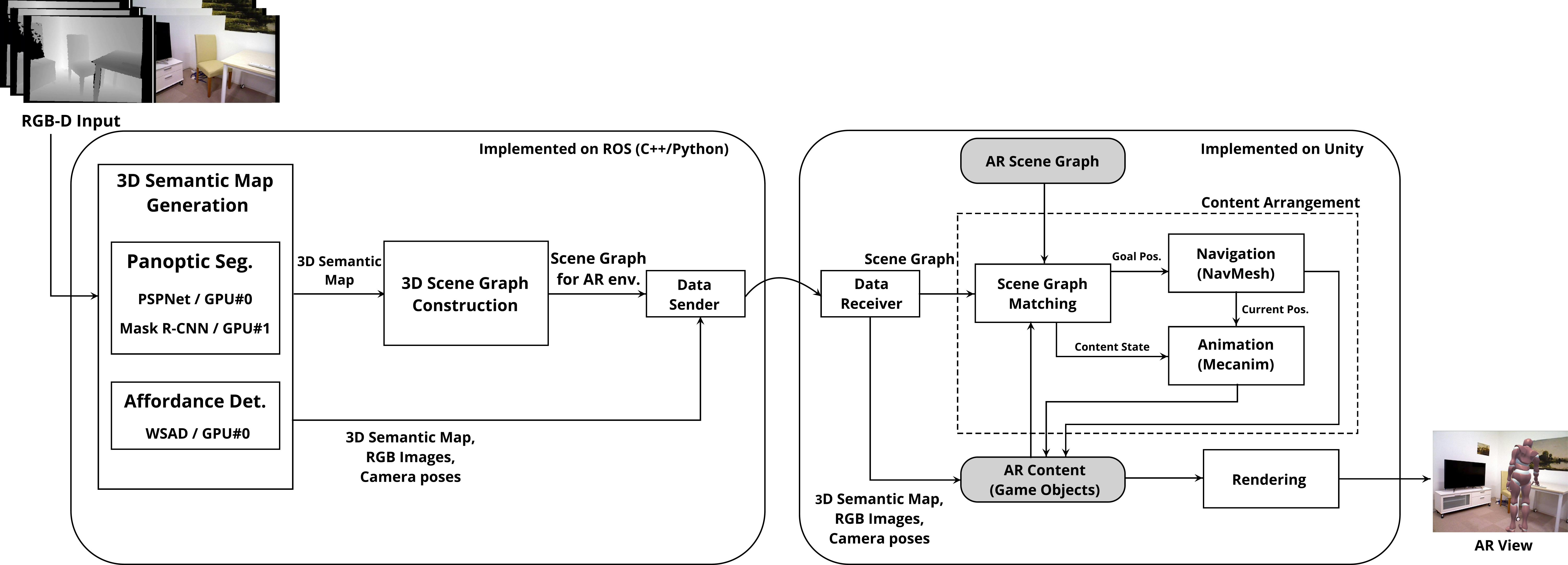}
  \caption{Configuration of prototype system.}
  \label{fig:prototyping_config}
\end{figure*}

\paragraph{\it{Content Arrangement.}}
A 3D scene graph $G$, representing a real AR environment in an abstract way, is incrementally updated according to the 3D semantic map update.
Whenever $G$ is updated, the content arrangement module compares the structure of $G$ with AR scene graph $G'$.
For the comparison, we use graph $G''$ derived from $G'$, which a node $v'_{c}$ corresponding to a virtual content, and its connected edges with relation $r_\mathrm{act}$ are removed.
When an obtained graph $G$ for an AR scene includes $G''$, that is, the context-representing graph $G''$ becomes a subgraph of the scene graph $G$ extracted from the real environment, it enables us to identify objects for AR in the 3D semantic map from the graph correspondences.
Thus we can obtain the content placement in a context-aware way that satisfies the assumed context shown in $G'$.
% AR体験シーンを抽象的に表現したシーングラフ$G$は意味情報マップの更新に従って逐次的に更新されるため，
% コンテンツ配置モジュールは$G$が更新される度に，ARシーングラフ$G'$との構造比較を行う．
% 構造比較は$G'$からARコンテンツに対応するノード$v'_c$及びそれらに接続された関係$r_\mathrm{act}$に対応するエッジを除去したグラフ$G''=(V'', E'')$を用いて行う．
% $G''$を含むようなグラフが現実シーンにおいて取得できるとき，
% つまり，ARコンテンツのシーングラフ$G''$が現実シーンのシーングラフ$G$の部分グラフとなるとき，
% $G$と$G''$の対応関係から意味情報マップで対応する各物体を特定できる．
% よって，想定したコンテキストのシーングラフ$G'$を満たすようにコンテンツを制御することで現実シーンに配置できる．

\section{Implementation of a Prototype System}
\label{sec:impl}
To verify the effectiveness of our framework, we constructed prototype system (\autoref{fig:prototyping_config}) that consists of two PCs: (1) A PC for semantic map generation and 3D scene graph construction with the input RGB-D streams; (2) a PC for AR content arrangement with the semantic map and input 3D scene graph.
RGB-D sequences are captured by off-the-shelf RGB-D sensors, in our case ASUS Xtion Pro Live, and camera poses are known or are estimated by Visual SLAM.
% Todo: Azure Kinectを例として挙げる必要があるか？
% 我々の提案する新たなARフレームワークの有効性を検証するために構築したプロトタイプシステムについて述べる．
% 本システムは\autoref{fig:prototyping_config}のように(1)RGB-D画像系列を入力とする意味情報マップ・3Dシーングラフ生成用のPCと，(2)意味情報マップと3Dシーングラフの情報を入力とするARコンテンツ配置用のPCから構成される．
% RGB-D画像系列はASUS Xtion Pro LIVEやAzure Kinectのような市販のRGBDカメラで取得され，カメラポーズは既知もしくはVisual SLAMにより推定されるものとする．

Here, we clarify the details of the settings for the learning in each recognition module included in the semantic map generation module.
We used PSPNet \cite{zhao2017pyramid} and Mask R-CNN\cite{he2017mask} as head CNNs for 2D semantic and instance segmentation, the same as in the original PanopticFusion \cite{narita2019panopticfusion}.
PSPNet was fine-tuned with ScanNet v2\cite{dai2017scannet} using pre-trained weights for ADE20K\cite{zhou2017scene}.
Mask R-CNN used different recognition categories for quantitative evaluation of the semantic mapping module and subjective evaluation by participants.
For quantitative evaluation, it was fine-tuned with ScanNet v2\cite{dai2017scannet} using pre-trained weights for COCO \cite{lin2014microsoft}.
In contrast, for subjective evaluation, we added \textit{tv} and \textit{remote} as additional categories to ScanNet's 20 categories, for the purpose of enhancing the AR experience.
To this end, we established and utilized the ScanNet20+2 dataset by extracting 4547 images including TVs and remote controls from COCO\cite{lin2014microsoft} and merged them with ScanNet's 17161 images.
The 80 categories of COCO were converted into ScanNet20+2 categories following \autoref{tab:scan_coco_table}.
We used the same pre-trained weights and parameters for training described in the original paper \cite{narita2019panopticfusion}.
% 意味情報マップ生成モジュールに含まれる各認識器の学習詳細について述べる．
% 我々はオリジナルのPanopticFusion\cite{narita2019panopticfusion}と同様にPSPNet\cite{zhao2017pyramid}とMask R-CNN\cite{he2017mask}をそれぞれ2D semantic and instance segmentationのためのCNNとして利用する．
% PSPNetはADE20K\cite{zhou2017scene}で事前学習した重みをScanNet v2\cite{dai2017scannet}を用いてfine-tuningした．
% Mask R-CNNについては，後述のsemantic mappingの定量評価と主観評価実験とで異なる認識カテゴリを学習した．
% 定量評価用にはCOCOデータセット\cite{lin2014microsoft}を事前学習した重みをScanNet v2\cite{dai2017scannet}を用いてfine-tuningした重みを用い，
% 主観評価用には，ARアプリの表現力を上げることを目的として，ScanNetの20カテゴリにTV, remoteの2カテゴリを追加して学習した．
% そのため，COCOデータセット\cite{lin2014microsoft}からTV, remoteを含む4,547枚を抽出し，ScanNetの画像17,161枚と統合したScanNet20+2データセットを構築して学習に用いた．
% なお，COCOの80カテゴリは\autoref{tab:scan_coco_table}にしたがってScanNet20+2クラスのカテゴリに変換した．
% 事前学習重みや学習時のパラメータ設定はいずれもPanopticFusion\cite{narita2019panopticfusion}に記載されているものと同様の設定を利用した．

As described above, we used the fully supervised settings of WSAD \cite{sawatzky2017weakly} as the affordance detection module and trained it on the ADE20K Affordance dataset we built by converting part labels of ADE20K\cite{zhou2017scene} into affordance labels for indoor scenes.
The ADE20K Affordance dataset consists of 3388 images (2980 training images, 408 test images). We extracted indoor scene images including one or more regions converted to the defined affordance labels according to \autoref{tab:aff_conv_table} and cropped them to $321 \times 321$ pixels according to an open implementation\cite{sawatzky2017weakly}.
The parameters for training were the same as the open implementation\cite{sawatzky2017weakly}.
In the evaluations, we performed the whole process of the semantic map generation and 3D scene graph generation with a PC equipped with an Intel Core i7-7800X CPU running at 3.6GHz and two NVIDIA GeForce GTX1080Ti GPUs.  
% アフォーダンス認識モジュールは前述の通り，ADE20K\cite{zhou2017scene}のパーツラベルをアフォーダンスラベルに変換して独自作成したADE20K Affordanceデータセットを用いてWSAD\cite{sawatzky2017weakly}の教師あり設定で学習させた．
% ADE20K Affordanceデータセットは，\autoref{tab:aff_conv_table}に従って所望のアフォーダンスラベルに変換された領域を1つ以上含む屋内シーン画像を抽出後，公開実装\cite{sawatzky2017weakly}に合わせて検出された領域を$321 \times 321$に切り抜いた画像3388枚(訓練2980枚, テスト408枚)で構成される．
% 学習時のパラメータは公開実装\cite{sawatzky2017weakly}と同じものを利用した．
% 意味情報マップ生成・シーングラフ生成処理はIntel Core i9-9900K CPU at 3.6GHzと2つのNVIDIA GeForce GTX1080Ti GPUsを備えたPCで実施した．

A generated semantic map and a 3D scene graph, together with RGB images and camera poses for AR, were sent in the form ROS messages to the PC for AR content arrangement via our modified rosbridge-based transmission module.
After receving the data from the PC for semantic map generation and 3D scene graph construction, the PC for AR content arrangement compared the AR scene graph with the 3D scene graph to control the AR contents. 
In particular, we used built-in functions in Unity to animate the AR character and to navigate the character on the \textit{floor} area of the semantic map. 
The PC for AR content arrangement was equipped with an Intel Core i7-7800X CPU running at 3.5GHz and an NVIDIA GeForce GTX1080 Ti GPU.
This configuration example is the same as the one used in the qualitative evaluations and subjective evaluation described later.
% 意味情報マップと3Dシーングラフ及びARのためRGB画像やカメラポーズは，独自に改良したrosbridgeベースの通信モジュールを介してROSメッセージをARコンテンツ配置用PCに伝送される．
% ARコンテンツ配置用PCでは，意味情報マップ・3Dシーングラフ生成用PCがデータを受信後，ARテンプレートと3Dシーングラフを逐次照合することでARコンテンツを制御する．
% 特に，ARキャラクタのアニメーションや意味情報マップ中の"Floor"領域上での移動経路探索には，Unityにビルトインされている機能を利用した．
% なお，ARコンテンツ配置用PCはIntel Core i7-7800X CPU at 3.5GHzとNVIDIA GeForce GTX1080 Tiを備えたものを利用した．
% この構成例は後述する定性評価，主観評価実験で用いるものと同じものである．

\begin{table}
  \caption{Conversion table of ScanNet20+2 from COCO80.}
  \label{tab:scan_coco_table}
  \scriptsize%
  \centering%
  \scalebox{0.7}[0.7]{
\begin{tabular}{|l|l|}\hline
  \textbf{ScanNet20+2} & \textbf{COCO80} \\ \hline
  bed             & bed \\ 
  chair          & chair  \\ 
  sofa             & coach \\ 
  table             & dining table \\ 
  refrigerator             & refrigerator \\ 
  toilet             & toilet \\ 
  sink             & sink \\ 
  tv             & tv \\ 
  otherfurniture             & bench \\ 
  remote             & remote \\ 
  background, otherprops             & other categories not shown on the table \\ \hline
  \end{tabular}
  }
\end{table}

\begin{table}
  \caption{Conversion table of ADE20K parts to affordances}
  \label{tab:aff_conv_table}
  \scriptsize%
  \centering%
  \scalebox{0.7}[0.7]{
\begin{tabular}{|l|l|}\hline
  \textbf{Affordance} & \textbf{ADE20K Parts}                                                                                                  \\ \hline
  sittable             & seat, seat\_cushion                                                                                                    \\
  supportable          & \begin{tabular}[c]{@{}l@{}}bed/base, bed/bedspring, shelf, countertop, work\_surface\end{tabular}                   \\
  pushable             & \begin{tabular}[c]{@{}l@{}}button(s), switch, light\_switch(es), electric(al)\_switch, push\_button(s)\end{tabular} \\
  openable             & knob, handle                                                                                                           \\
  leanable             & \begin{tabular}[c]{@{}l@{}}chair/back, chair/back\_pillow, sofa/back, sofa/back\_pillow\end{tabular} \\ \hline
  \end{tabular}
  }
\end{table}

% \begin{table}[t]
%   \caption{Conversion Table of ADE20K parts to affordance}
%   \label{tab:aff_conv_table}
%   \scriptsize%
% 	\centering%
%   \begin{tabu}{|l|l|}\hline
%     \textbf{Affordance} & \textbf{ADE20K Parts} \\ \hline
%     sittable & seat, seat\_cushion \\ 
%   supportable & \begin{tabular}[c]{@{}l@{}}bed/base, bed/bedspring, shelf, countertop, \\ work\_surface \end{tabular}\\ 
%     pushable & \\
%     openable & \\
%     leanable & \\
%   \end{tabu}
% \end{table}

\section{Experiments}
\subsection{Evaluation of Semantic Mapping}
To evaluate the semantic mapping performance of our prototype system quantitatively, we performed experiments on both the 3D panoptic segmentation task and the affordance detection task with the system.
% プロトタイプシステムでのsemantic mapping性能の定量評価のため，3D panoptic segmentationタスクとaffordance detectionタスクのそれぞれについて実験を行った．

For the 3D panoptic segmentation task, we employed 3D panoptic quality (PQ) in the ScanNet dataset as a metric and compared our system with PanopticFusion\cite{narita2019panopticfusion} without CRF (\autoref{tab:pq_eval}). PQ is defined as the multiplication of recognition quality (RQ) and segmentation quality (RQ) \cite{kirillov2019panoptic}.
\autoref{tab:pq_eval} shows that there was little difference in the PQ/RQ/SQ scores between PanopticFusion and our system.
Therefore, this shows that the proposed framework can integrate an additional recognition module such as an affordance detector without sacrificing its object recognition performance.% 3D panoptic segmentationタスクでは，ScanNetデータセットでの3D panoptic qualityを指標としてCRFなしのPanopticFusion\cite{narita2019panopticfusion}と比較した．
% Table \ref{tab:pq_eval}に示す通り，PanopticFusionと我々のsemantic mappingとで，各カテゴリ・カテゴリ平均，PQ/SQ/RQそれぞれについて性能に大きな差はないことが確認できる．
% 從って提案フレームワークでは，物体認識性能を犠牲にすることなく，アフォーダンス等の追加の認識機能を統合することが可能である．

\begin{figure}[t]
  \setlength\abovecaptionskip{0pt}
  \centering 
  \includegraphics[width=\columnwidth]{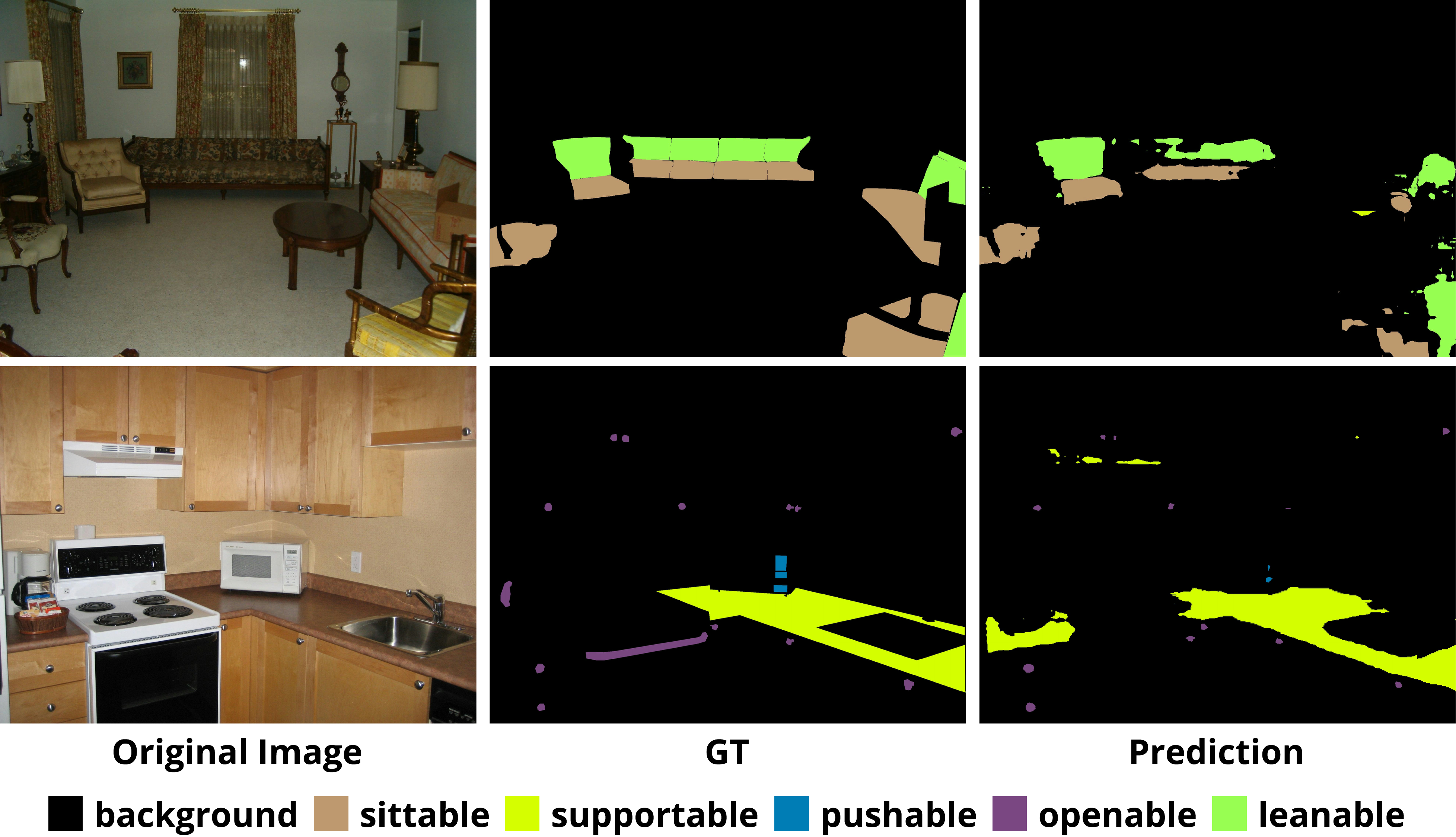}
  \caption{Examples of affordance detection results.}
  \label{fig:affordance_detection}
\end{figure}

In the affordance detection task, we evaluated the affordance segmentation results with pixel accuracy and mean Intersection-over-Union (mIoU) on test images of the ADE20K Affordance dataset introduced in \autoref{sec:impl}.
The pixel accuracy for the overall image was 64.0 \%, whereas the average score of classes turned out to be 44.4 \%.
These results reflected the fact that the regions with \textit{pushable} or \textit{openable} labels, corresponding to parts such as buttons and knobs, were so small (\autoref{fig:affordance_example}) that they were more challenging to detect than others.
Other labels, \textit{sittable}, \textit{leanable} and \textit{supportable} obtained good estimation results with accuracies over 60 \%.
\autoref{fig:affordance_detection} shows examples of detection results for affordance masks.
% affordance detectionタスクでは，\ref{sec:impl}節記載のADE20K Affordanceデータセットのテスト画像を基にpixel accurayとmIoU(mean Intersection-over-Union)を指標とした評価を行った．
% pixel accurayについて，全画素のスコア64.0％に対してクラス平均は44.4％であった
% これはボタンやドアノブ等部位を表すpushableやopenableラベルを持つ領域が小領域（\ref{fig:affordance_example}）であり，検出が比較的困難であることを反映している．
% また，その他のsittable, leanable, supportableラベルについては60％を超えるaccurayで良好に推定されている．
% 図\ref{fig:affordance_detection}にアフォーダンスマスクの検出結果例を示す．

\begin{table}[t]
  \renewcommand{\baselinestretch}{0.8}
  \caption{3D panoptic segmentation results on ScanNet (v2) open test set.}
  \label{tab:pq_eval}
  \centering
  \scalebox{0.7}[0.7]{
     \begin{tabular}{c|c||c|cc}
                                                                                                                        \hline
                                                                                                                method & metric & all & \textit{things} & \textit{stuff} \\ \hline
      \multirow{3}{*}{\begin{tabular}[c]{@{}c@{}}PanopticFusion\\ w/o CRF\cite{narita2019panopticfusion}\end{tabular}} & PQ & 29.7 & 26.7 & 56.7 \\
                                                                                                                       & SQ & 71.2 & 71.4 & 69.5 \\
                                                                                                                       & RQ & 41.1 & 36.8 & 79.6 \\ \hline
      \multirow{3}{*}{\begin{tabular}[c]{@{}c@{}}PanopticFusion \\ w/ multi-label integration \\ (our prototype system)\end{tabular}}                                               & PQ & 30.5 & 27.5 & 56.9 \\
                                                                                                                       & SQ & 71.3 & 71.4 & 70.1 \\
                                                                                                                       & RQ & 42.2 & 38.1 & 79.2 \\ \hline
     \end{tabular}
  }
\end{table}

\begin{table}
  \renewcommand{\baselinestretch}{0.8}
  \caption{Pixel accuracy and mean IoU on ADE20K Affordance dataset.}
  \label{tab:aff_eval}
  % \scriptsize%
	\centering%
  \scalebox{0.7}[0.7]{
    \begin{tabular}{c|c*{5}{c}||c}
    % \toprule
    \hline
    \multicolumn{7}{c||}{pixel accuracy (\%)}                                & \multirow{2}{*}{mIoU(\%)} \\ %\cline{1-7}
    overall & avg & sittable & supportable & pushable & openable & leanable &                       \\ \hline
    % \midrule
    64.0 & 44.4 & 67.0 & 62.0 & 7.0 & 17.0 & 69.0 & 44.2 \\ \hline
    % \bottomrule
    \end{tabular}%
  }
\end{table}

\subsection{Evaluation of Scene Object Abstraction}
We quantitatively evaluated the detection performance of OBBs using the Scan2CAD dataset \cite{Avetisyan2019scan2cad} built for CAD model retrieval and alignment tasks.
The Scan2CAD dataset contains information associating similar 3D CAD models of ShapeNet \cite{angel2015shapenet} with real scene 3D scans of ScanNet via 3D keypoint correspondences and also includes annotations for 3D OBBs of CAD models aligned with the scans.

Setting the annotations of 3D OBBs as the ground truth, we evaluated median orientation errors of the predicted OBBs and mIoU/precision/recall for True Positive detection results.
In our evaluation process, panoptic and affordance labels were estimated online and mapped on the ground truth geometries of ScanNet, which yielded 3D semantic maps for 3D scene graph construction.
To associate categories defined in Scan2CAD with those of ScanNet20+2 described in \autoref{sec:impl}, we treated the \textit{display} category of Scan2CAD as the \textit{TV} category of ScanNet20+2, and excluded \textit{trashbin}/\textit{other}, which do not have suitable counterpart categories, from the evaluation targets.
In addition, we used z-axis aligned poses for the 3D OBB ground truth.
% 提案フレームワークで生成する3D scene graphは，3D OBBとして抽象化した各物体のカテゴリや位置，大きさ，向きの検出性能に大きな影響を受ける．
% そこで各物体のOBBとしての検出性能を，CADモデルの検索・アラインメントタスクのデータセットであるScan2CADデータセット\cite{Avetisyan2019scan2cad}を利用して定量的に評価した．
% Scan2CADデータセットは，ScanNetデータセットと同じ現実シーンの3Dスキャンに対して，
% ShapeNet\cite{angel2015shapenet}からの類似CADモデルを3Dキーポイントで対応付けた情報を含み，更にアラインメントされたCADモデルに対する3D OBBのアノテーションを持つ．
% この3D OBBアノテーションを真値として，プロトタイプシステムで推定したOBBの方向誤差中央値，True Positiveな検出結果に対するmIoU及びprecision, recallを評価した．
% なお本評価では，プロトタイプシステムによってオンラインで推定したpanopticラベルマップとaffordanceラベルマップに，ScanNetの真値シーン形状を対応付けることで3D意味情報マップを取得した．
% また，\ref{sec:impl}節で述べたScanNet20+2カテゴリとScan2CADで定義されたカテゴリを対応付けるため，Scan2CADのdisplayカテゴリはScanNet20+2カテゴリのtvカテゴリとして扱い，
% ScanNet20+2カテゴリに対応するカテゴリの存在しないtrashbinやotherは評価対象外とした．
% 加えて，3D OBB真値の姿勢については，鉛直軸周りの回転のみに限定した姿勢を利用した．

\autoref{tab:obb_angle_eval} shows median orientation errors of successfully detected OBBs. They were in the same category as the ground truth and had IoUs of 0.25/0.5 or more. 
The results confirmed that the proposed system could estimate the orientations of the OBBs with errors of several degrees for the majority of OBBs.
% 真値と推定OBB間で同一カテゴリかつIoUが0.25または0.5以上の場合に推定OBBを検出成功とし，検出が成功したOBBについて方向誤差中央値をTable \ref{tab:obb_angle_eval}に示す．
% 検出が成功したOBBについては，過半数のOBBについて数度程度の誤差でOBBの向きが推定できていることが確認できた．

We defined True Positive OBBs as successfully detected OBBs with orientation errors of 20 degrees or less and calculated the TP IoU/precision/recall, as shown in \autoref{tab:obb_tp_eval}.
The results show that \textit{sofa} was recognized particularly well. It suggests that the orientation estimation by the affordance and the large object size improved its detection performance.
% また，真値と推定OBBのカテゴリが同じかつ，IoUが0.5または0.25以上，方向誤差が20度以内の場合をTrue PositiveなOBB検出結果とした，TP(True Positive) IoU，presicion, recallをTable \ref{tab:obb_tp_eval}に示す．
% アフォーダンスによる方向推定を行い，なおかつ物体サイズの大きいsofaは特に良好に認識されていることが確認できる．

\begin{figure}[t]
  \setlength\abovecaptionskip{0pt}
  \centering 
  \includegraphics[width=\columnwidth]{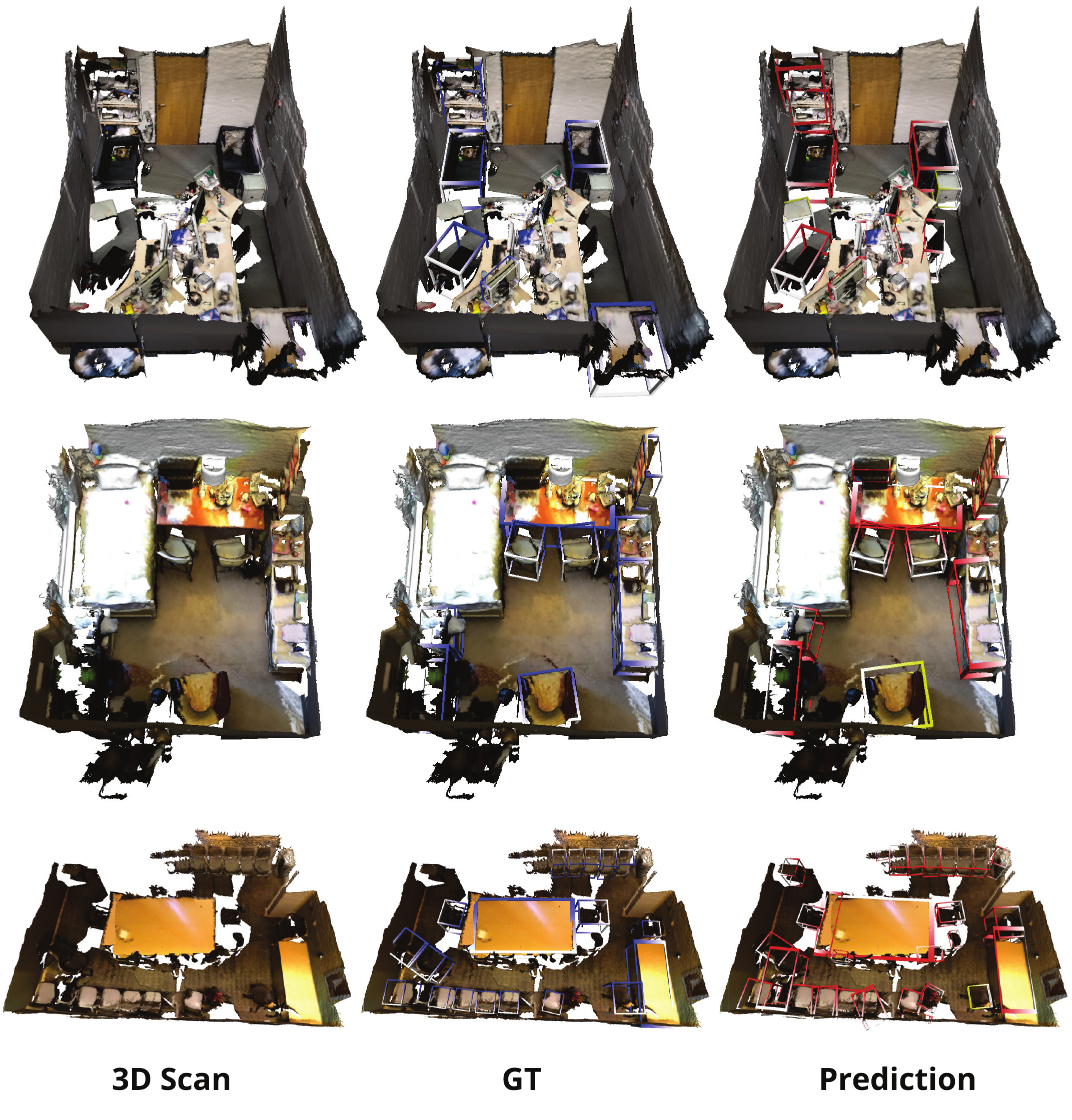}
  \caption{Examples of predicted OBBs with the ground truths. 
           Blue and red boxes are ground truth and predicted OBBs, respectively.
           OBBs without explicit orientation are colored yellow. 
           Color gradient (white to blue/red) means the forward direction of the OBBs.
           (top:scene0025\_02, middle:scene0144\_00, bottom:scene0414\_00)}
  \label{fig:obb_examples}
\end{figure}

Although the orientations of \textit{chair} were similarly estimated by using the affordance, it can be inferred that due to the lack of shape information in some parts of the chair (\autoref{fig:obb_examples} bottom-right), which resulted from incomplete scanning, its IoU decreased, resulting in a lower Recall rate.
Both \textit{sofa} and \textit{chair} in particular showed better Precision rates, which means that their orientations were predicted more accurately by using affordances rather than other categories.
\autoref{fig:obb_examples} shows examples of detected OBBs in the ScanNet scenes.
% 同様にアフォーダンスによる方向推定を行うchairについては，十分に計測されないことによる欠損によって（図\ref{fig:obb_examples} 中段右）IoUが低下し，Recallレートが低くなっていると推察できる．
% この両者についてはアフォーダンスによって，他のカテゴリに比べて良好に方向を推定できることからPrecisionが特に高いことが分かる．
% 図\ref{fig:obb_examples}にOBBの検出結果例を示す．

\begin{table}[t]
  \caption{Median orientation error in Scan2CAD dataset.}
  \label{tab:obb_angle_eval}
  % \scriptsize%
	\centering%
  \scalebox{0.7}[0.7]{
    \begin{tabular}{l|c|*{9}{c}}
    \hline
                                                                                            & IoU th. & bath & bkshf & cab  & chair & tv   & sofa & tabl \\ \hline
  \multirow{2}{*}{\begin{tabular}[c]{@{}c@{}}Median orientation\\error [deg.]\end{tabular}} & 0.25    & 2.16 & 2.24  & 2.10 & 4.01  & 3.10 & 1.82 & 1.55 \\ 
                                                                                            & 0.50    & 1.21 & 1.94  & 1.25 & 2.74  & 1.87 & 1.78 & 1.31 \\ \hline

    \end{tabular}%
  }
\end{table}

\begin{table}[t]
  \caption{TP IoU, precision, and recall evaluation in Scan2CAD dataset.}
  \label{tab:obb_tp_eval}
  % \scriptsize%
	\centering%
  \scalebox{0.7}[0.7]{
    \begin{tabular}{l|c|*{9}{c}}
    \hline
                                                                         & IoU th. & bath  & bkshf  & cab   & chair  & tv    & sofa  & tabl  & avg   \\ \hline
  \multirow{2}{*}{\begin{tabular}[c]{@{}c@{}}TP IoU(\%)\end{tabular}}    & 0.25    & 53.97 & 50.88  & 58.43 & 51.46  & 46.45 & 66.59 & 55.38 & 54.74 \\ 
                                                                         & 0.50    & 63.73 & 61.36  & 70.06 & 63.38  & 56.74 & 68.44 & 65.93 & 64.23 \\ \hline
  \multirow{2}{*}{\begin{tabular}[c]{@{}c@{}}Precision(\%)\end{tabular}} & 0.25    & 50.00 & 44.07  & 15.49 & 68.14  & 39.46 & 72.00 & 32.39 & 45.94 \\ 
                                                                         & 0.50    & 28.95 & 23.73  & 10.20 & 31.59  & 15.68 & 66.67 & 20.45 & 28.18 \\ \hline
  \multirow{2}{*}{\begin{tabular}[c]{@{}c@{}}Recall(\%)\end{tabular}}    & 0.25    & 31.67 & 36.79  & 30.38 & 45.20  & 38.22 & 47.79 & 28.93 & 37.00 \\ 
                                                                         & 0.50    & 18.33 & 19.81  & 20.00 & 20.95  & 15.18 & 44.25 & 18.26 & 22.40 \\ \hline
  \end{tabular}%
  }
\end{table}

\subsection{Subjective Evaluation of Context-Aware Content Arrangement}
% Unfortunately, to the best of our knowledge, there is no dataset available for direct quantitative evaluation of the accuracy of 3D scene graph construction.
% Instead, 
To evaluate whether or not the proposed framework was able to obtain context-aware content arrangements using 3D scene graphs, we performed a subjective evaluation of AR content arrangements in various scenes.
% 我々が知る限りにおいて，残念ながら3Dシーングラフの生成精度を直接的かつ定量的に評価可能なデータセットは存在しない．
% そこで我々は3Dシーングラフを直接評価するのではなく，提案フレームワークの目的であるコンテキストに応じたコンテンツ配置が生成した3Dシーングラフを用いて実現出来ているかを主観評価実験により確認した．

First, we set an AR scene graph $G'$.
Then, from RGB-D data and camera poses of the ScanNet dataset, we extracted a 3D semantic map and a 3D scene graph $G$ by using the prototype system.
For a scene satisfying the AR scene graph $G'$, we placed AR content on 3D scans based on the comparison result of $G$ and $G'$, achieving a context-aware arrangement (CA).
Apart from that arrangement procedure, we performed random arrangement (RA) of the AR content so that the categories of objects with which the virtual content interacted matched with those of $G'$.
We presented the arrangement results in the same scene from the two approaches above for the AR scene graph $G'$ to participants. The participants gave subjective evaluations for the set context-awareness of the arrangement result by using a five-point Likert scale: from "strongly unexpressed" to "strongly expressed".
As participants, we selected 12 people who did not have much knowledge of the proposed framework.  Note that the participants were not informed of the ways of arranging contents for the two results, and the order of the presented contents was randomized.

% 具体的には以下の条件にて評価実験を実施した．
% まず，表現したいARシーングラフ$G'$を設定する．
% 次に，プロトタイプシステムによってScanNetデータセットのRGB-Dデータ及びカメラポーズからシーンの3D意味情報マップと3Dシーングラフ$G$を生成する．
% そして，ARシーングラフ$G'$を満足するシーンに対して，3Dシーングラフ$G$と$G'$の照合結果に基づいて3Dスキャン中にコンテンツ配置を行う．
% そのコンテンツ配置とは別に，ARシーングラフ$G'$中の仮想コンテンツのインタラクション対象の物体カテゴリのみ一致させるランダムなコンテンツを配置（以下，ランダム配置）を行う．
% 被験者には，ARシーングラフ$G'$に対する上記2種類のコンテンツ配置を同一のシーンに対して適用した結果を提示し，
% 「全く表現していない」から「よく表現している」までのfive-point lickertスケールにより設定されたコンテキストの表現度合いを主観的に評価する．
% 被験者は，提案フレームワークに関する深い知識を持たない12名であり，2種類のコンテンツ配置方法についても情報は与えられず，
% 提示されるコンテンツ配置の順序はランダムに決定されている．

\begin{figure}[t]
  \setlength\abovecaptionskip{0pt}
  \centering 
  \includegraphics[width=\columnwidth]{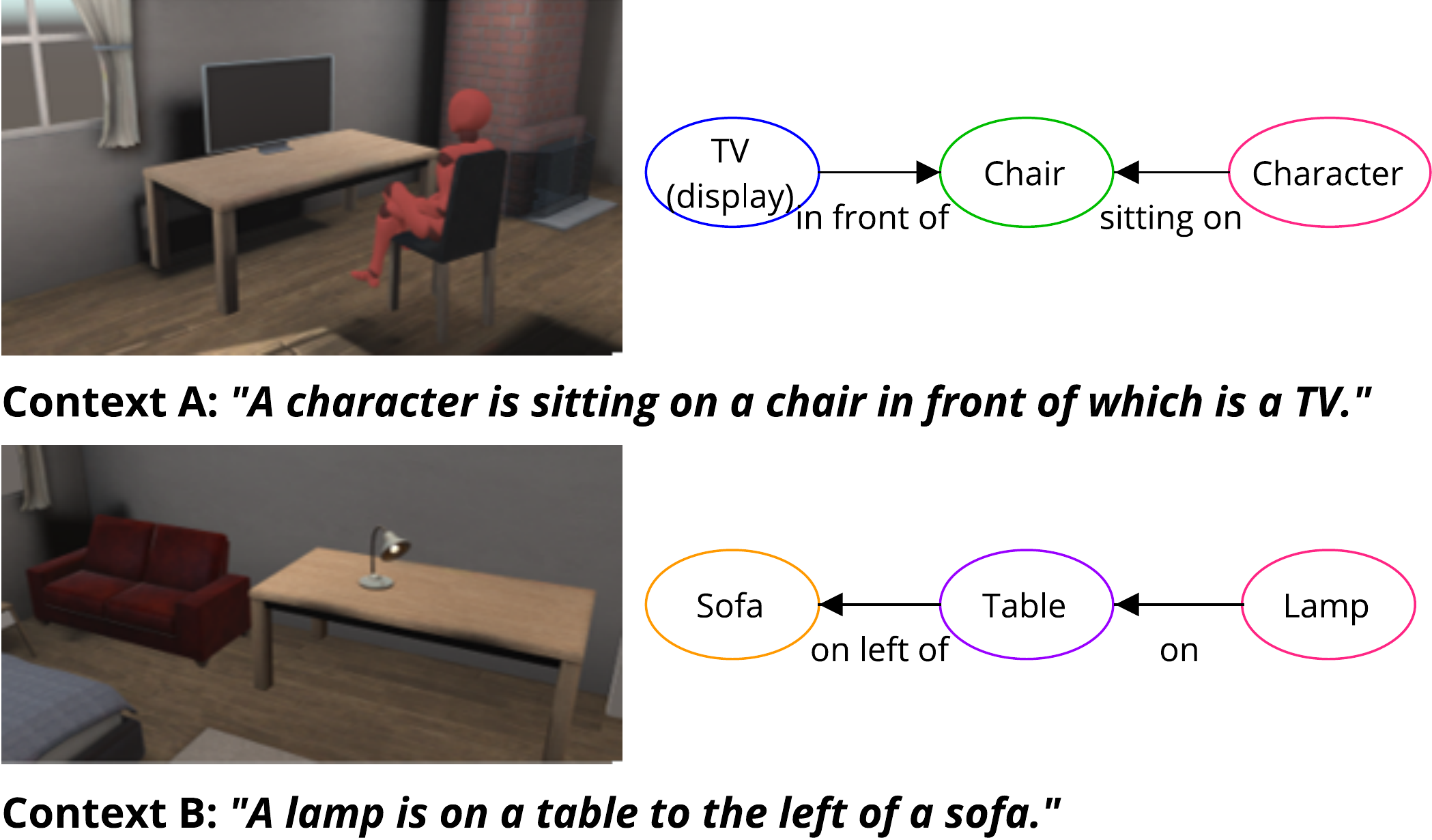}
  \caption{Contexts for subjective evaluation. Scene examples (left) and the scene graphs (right) representing the contexts were provided to subjects.}
  \label{fig:context_for_subjective_evaluation}
\end{figure}

In our experiments, we set the following two types of scene graphs as $G'$ (\autoref{fig:context_for_subjective_evaluation}):
\begin{description}
  \item[Context A] \textit{"A character is sitting on a chair in front of which is a TV."} (\autoref{fig:context_for_subjective_evaluation} top)
  \item[Context B] \textit{"A lamp is on a table to the left of a sofa."} (\autoref{fig:context_for_subjective_evaluation} bottom)
\end{description}

To clarify contexts provided by $G'$, we presented the above text descriptions that describe a scene graph in natural language to participants, together with $G'$.
% また，シーングラフの読み方による誤解を防ぐことを目的として，被験者には設定したシーングラフ$G'$と共にそのシーングラフを自然言語で表現した上記と同様のテキストも提示した．

There were a total of 33 scenes in the ScanNet dataset satisfying the contexts above: 27 for Context A and 6 for Context B.
% Todo: ScanNetの膨大なシーンのうちからの33シーンというニュアンスが抜けていた
For these scenes, we performed both CA by the prototype system and RA using the categories of interaction objects. The arrangement results were presented to the participants for evaluation.
% また，シーングラフの読み方による誤解を防ぐことを目的として，被験者にはARシーングラフ$G'$と共にそのシーングラフを自然言語で表現した上記と同様のテキストも提示した．
% 上記コンテキストを満足するScanNetデータセット中のシーンは，Context Aについて27シーン，Context Bについて6シーンの合計33シーンであった．

\begin{figure}[t]
  \setlength\abovecaptionskip{0pt}
  \centering 
  \includegraphics[width=\columnwidth]{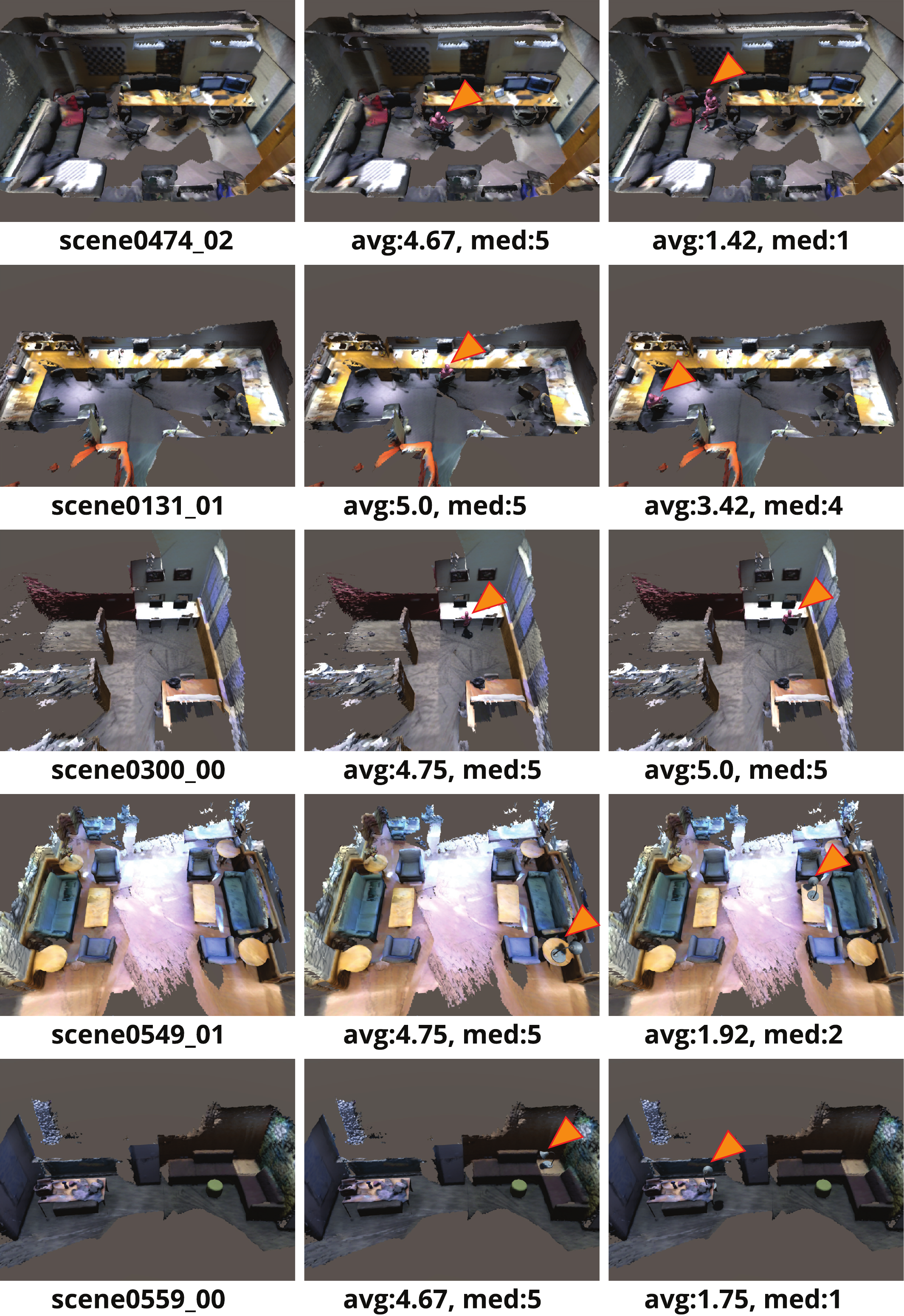}
  \caption{3D scans (left column), context-aware content arrrangement (middle column), and random content arrangement (right column) for subjective evaluation. 
  Average and median scores were estimated from subjects' responses for each scene.
  Orange triangles indicate locations of arranged contents.
  }
  \label{fig:scan_and_content_arrangement}
\end{figure}

\begin{figure}[t]
  \setlength\abovecaptionskip{0pt}
  \centering 
  \includegraphics[width=\columnwidth]{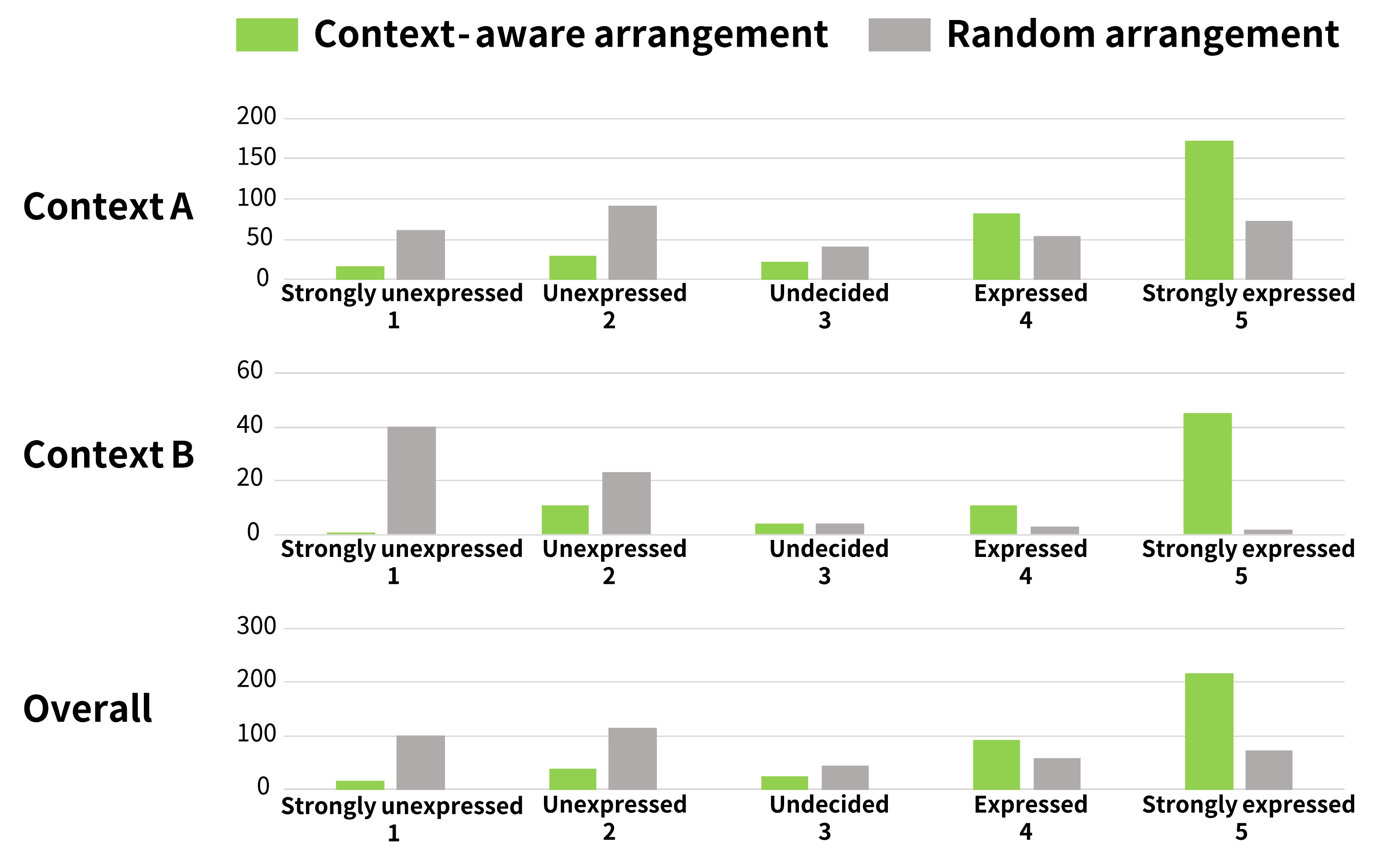}
  \caption{Evaluation scores from subjects for each context and overall, in context-aware and random arrangements.}
  \label{fig:subjective_evaluation_results}
\end{figure}

\autoref{fig:scan_and_content_arrangement} shows 3D scans obtained by the prototype system and example results for CA and RA on those scans.
The votes for the degrees of expression against each context and the whole result by CA/RA are shown in \autoref{fig:subjective_evaluation_results}.
In the following, we describe the evaluation results with the notations "Strongly unexpressed" as 1 point, and "Strongly expressed" as 5 points.
% 図\ref{fig:scan_and_content_arrangement}にプロトタイプシステムによる3Dスキャンとその3Dスキャンへのコンテキストに応じたコンテンツ配置及び，ランダム配置の結果例を示す．
% また，図\ref{fig:subjective_evaluation_results}にコンテキストに応じたコンテンツ配置とランダム配置における，各コンテキスト及び全体での表現度合いに対する被験者からの投票数を示す．
% なお以下では，「Strongly unexpressed」を1-point，「Strongly expressed」を5-pointとして評価結果について述べる．

For the results by CA, the rates of answers that the participants evaluated as context-expressed was 79\% for Context A, 78\% for B, and 79\% for the whole. 
In contrast, the results by RA were 39\% for Context A, 7\% for B, and 33\% for the whole.
In addition, we performed the Wilcoxon Rank Sum Test for the CA/RA results and confirmed that both the results for each scene and the total showed statistically significant differences (p-values were less than 0.01).
These findings verified that the CA, the context-aware content arrangement using the scene graph obtained by the proposed framework, was effective for expressing the contexts of AR contents in real environments.
% コンテキストに応じたコンテンツ配置について，コンテキストが表現されていた（4- or 5-point）と評価した回答率は，
% コンテキストA，コンテキストB，全体でそれぞれ，79％，78％，79％だった．
% 対して，ランダム配置については，コンテキストが表現されていた（4- or 5-point）と評価した回答率は，
% コンテキストA，コンテキストB，全体でそれぞれ，39％，7％，33％だった．
% また，コンテキストに応じた配置とランダム配置とでWilcoxon Rank Sum Testを行った結果，各コンテキストでも全体でも統計的有意差(p<0.01)が確認できた．
% 以上の結果から，提案フレームワークのシーングラフによるコンテキストに応じたコンテンツ配置は，設定したコンテキストを表現するのに有効であることを確認した．

Focusing on the evaluation of each scene, the CA achieved statistically significant evaluation (p-values were less than 0.05) in 19 scenes out of the total 33 and showed better degrees of expression than RA.
In many of such scenes, as shown in lines 1, 2, 4, and 5 in \autoref{fig:scan_and_content_arrangement}, CA can express the scene contexts (\autoref{fig:context_for_subjective_evaluation}) more appropriately than RA.
However, there was no significant difference in some scenes where the AR scene graph could be included by selecting any object to be interacted with in the scenes, as shown in line 3 in \autoref{fig:scan_and_content_arrangement}.
% 各シーンでの評価については，全33シーン中19シーンで統計的優位(p<0.05)にコンテキストに応じた配置の方が表現に優れる結果となった．
% その多くのシーンでは，図\ref{fig:scan_and_content_arrangement}中の1,2,4,5行目に示すコンテンツ配置のように，
% コンテキストに応じたコンテンツ配置の方がランダム配置に比べて，\ref{fig:context_for_subjective_evaluation}のシーンをより適切に表現できている．
% しかし，図\ref{fig:scan_and_content_arrangement}中の3行目のシーンのように，シーン中のインタラクション対象の物体のどれを選択してもARシーングラフを満足できるシーンでは，有意差は見られなかった．

\subsection{Online Context-Aware AR Demonstration}
\begin{figure}[t]
  \setlength\abovecaptionskip{0pt}
  \centering 
  \includegraphics[width=\columnwidth]{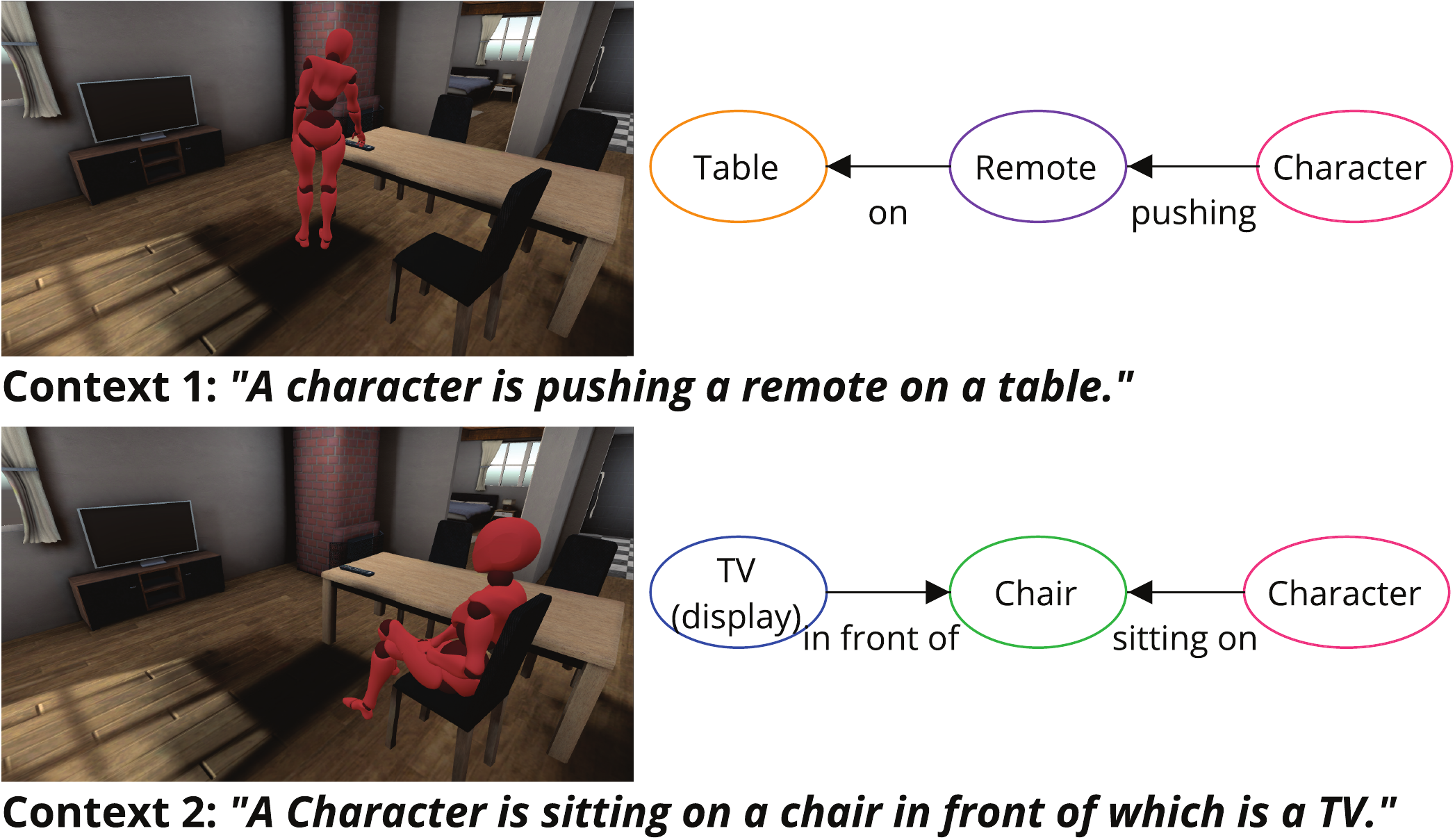}
  \caption{Contexts for online AR demonstration.}
  \label{fig:contexts_for_online_demo}
\end{figure}

We conducted experiments to demonstrate that our prototype system was capable of achieving an AR application in various indoor spaces depending on defined contexts.
In these demonstrations, we set the two contexts described in \autoref{fig:contexts_for_online_demo} in advance and verified whether each AR context could be expressed in three different real scenes.
Here, by applying Context 2 after Context 1 is expressed in the scenes, we expressed a series of interactions of a virtual character: a character "pushes a remote control on a table" to turn on the TV  and then "sits on a chair in front of which is a TV" to watch the TV.
As real environments, we employed an open space scene, a meeting room scene, and a living room scene with a TV, a table, a remote control and a chair. 
Note that our prototype system works without using absolute coordinates, and since it arranges AR contents based on the recognized 3D semantic map and scene graph, the AR program for the experiment did not need any setting change at all in each scene.
% 我々のプロトタイプシステムによって設定したコンテキストに応じたARアプリケーションを様々な空間で実現できること実証するための実験を行なった．
% この実証実験では，事前に図\ref{fig:contexts_for_online_demo}に示す2つのコンテキストを設定し，
% それらのコンテキストを3つの異なる現実シーンでAR表現できることを確認する．
% また，コンテキスト1が現実シーンで表現されたタイミングでコンテキスト2を適用することで，
% キャラクタが「テーブルの上のリモコンを押して」TVを電源ONした後，TVを見るために「TVを前にしたイスに座る」といった一連のキャラクタの行動を表現する．
% 現実シーンには，TV，テーブル，リモコン，イスのあるオープンスペースシーン，オフィスシーン，リビングシーンを利用した．
% なお我々のプロトタイプシステムでは，絶対座標に関する情報を全く利用せず，認識された3D意味情報マップ及びシーングラフを基にコンテンツを配置するため，
% シーン毎にARプログラム中の設定変更は全く必要ないことに注意されたい．

\autoref{fig:online_ar_demo} shows the online AR demonstrations in each scene (see the supplementary video for the details of each scene).
These demonstrations confirmed that, in each of the scenes, the proposed framework could recognize individual objects and their affordances, and spatial relations between objects in an online fashion.
In addition, they showed that the framework could obtain context-aware content arrangements based on the recognition results. 
\autoref{fig:scene_graph_matching_context} illustrates the scene graph recognized in the open space scene and its relationship between Context 1 and Context 2. 
\autoref{tbl:runtime_analysis} shows the runtime analysis for each component of the proposed framework in executing the AR application in the open space scene.
% 図\ref{fig:mapping_results}に各シーンにおけるオンラインARデモの様子を示す（各シーンの詳細は添付のデモビデオを参照のこと）．
% 各シーンにおいて，オンラインで個々の物体やアフォーダンスを認識し，更に物体間の関係を認識出来ていることが確認出来る．
% また，その認識結果を基に設定されたコンテキストに応じたコンテンツ配置が実現できることが確認出来る．
% リビングシーンで認識されたシーングラフと，コンテキスト1，2との関係を図\ref{fig:scene_graph_matching_context}に示す．
% 表\ref{tbl:runtime_analysis}にリビングシーンでのARアプリケーション実行時の提案フレームワークの各コンポーネントの計算時間を示す．

With this experiment, we demonstrated that the proposed method is capable of achieving AR expressions in various physical environments, according to the set contexts, and expressing a series of actions by switching the contexts.
We consider that the proposed framework will enable more storytelling, space-adaptive AR applications that go beyond conventional AR applications, like virtual contents appearing in the space randomly. 
% 本実験により，設定したコンテキスに応じたAR表現が様々な現実シーンにおいて実現可能であり，コンテキストを切替えることで一連の行動を表現する
% ことが可能であることを示した．
% 我々は提案フレームワークにより，従来よくあるようなコンテンツが空間中にランダムに出現するようなARアプリケーションから一線を画し，
% よりストーリー性のある空間適応的なARアプリケーションが実現可能であると考える．

\begin{figure*}[t]
  \centering 
  \includegraphics[width=2\columnwidth, height=8cm]{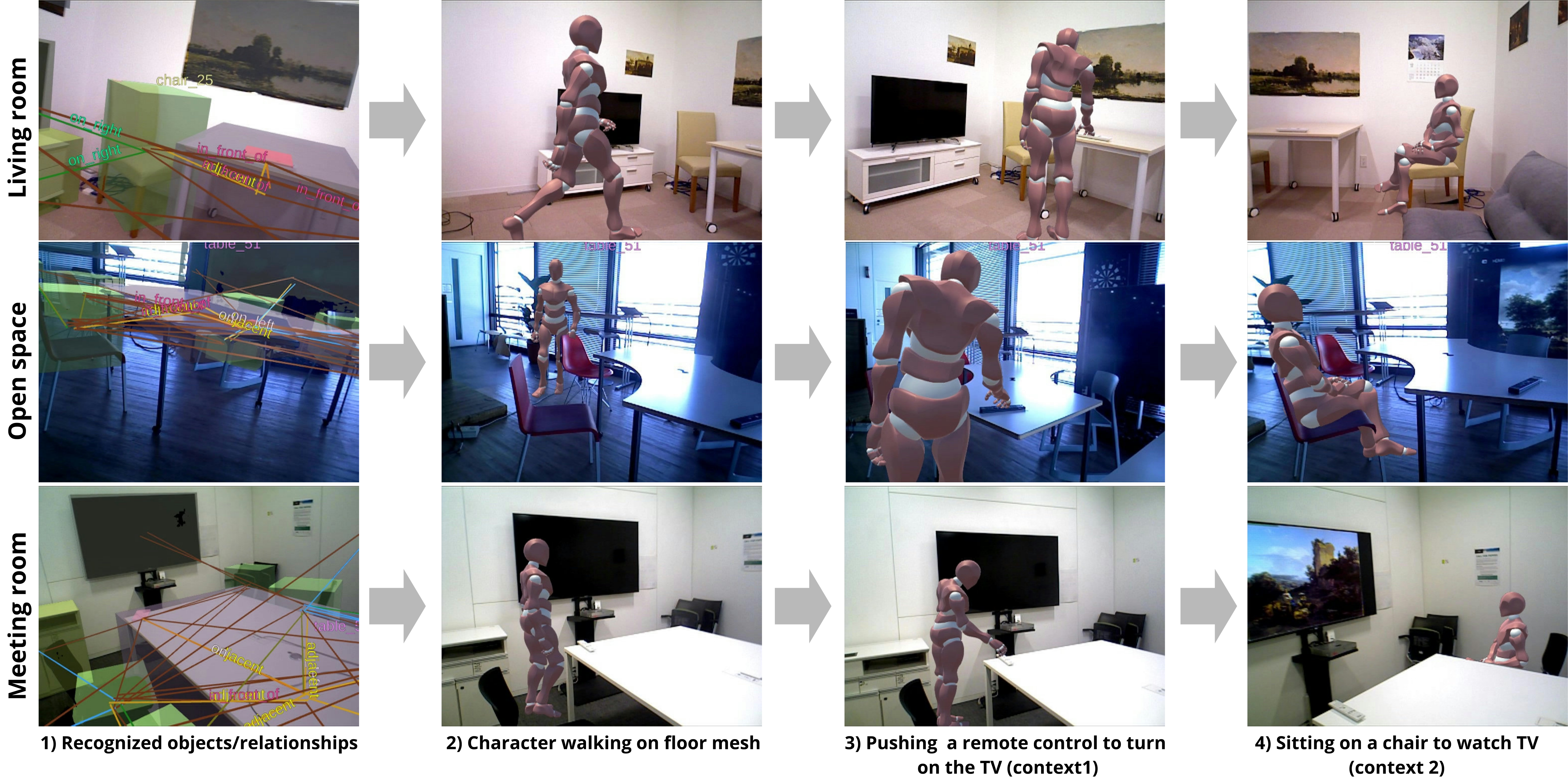}
  \caption{Online context-aware AR demo in various scenes.}
  \label{fig:online_ar_demo}
\end{figure*}

\begin{figure}[t]
  \centering 
  \includegraphics[width=\columnwidth]{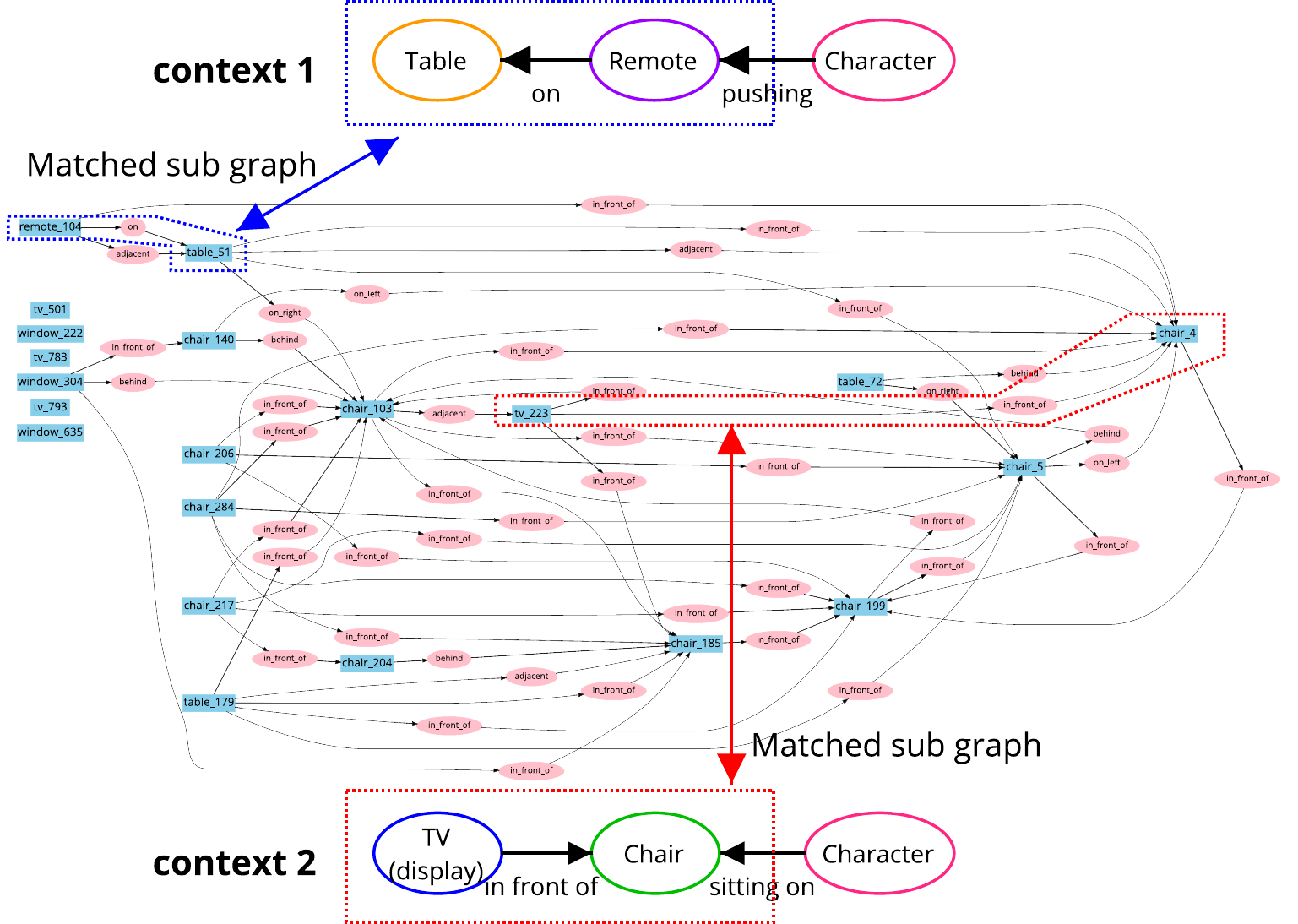}
  \caption{Matched sub graphs between generated scene graphs and AR scene graphs in the open space of \autoref{fig:online_ar_demo}. Note that "near" relationships are not visualized.}
  \label{fig:scene_graph_matching_context}
\end{figure}

\begin{table}[t]
  \renewcommand{\baselinestretch}{0.8}
  \caption{Run-time analysis of our prototype system running in the open space.}
  \label{tbl:runtime_analysis}
  \centering
  \scalebox{0.95}[0.95]{
     \begin{tabular}{llr}
        \hline
        \textbf{Frequency}      & \textbf{Component}            & \textbf{time} \\ \hline\hline
        Every RGB-D frames      & TSDF integration              & 103 ms \\ \hline
        Every Mask R-CNN frames & PSPNet                        & 124 ms \\
                                & Mask R-CNN                    & 266 ms \\
                                & Affordance detection          & 492 ms \\
                                & Panoptic label integration    & 58 ms \\ 
                                & Affordance label integration  & 56 ms \\ \cline{2-3} 
                                & Panoptic label tracking       & 6  ms \\
                                & Probability integration       & $\sim$1 ms \\ \hline
        Every 1 sec.            & Mesh extraction               & 14 ms \\ 
                                & Scene object abstraction      & 173 ms \\ 
                                & Scene graph construction      & $\sim$2 ms \\ \hline
        Triggered timing        & Content arrangement           & $\sim$1 ms \\ \hline
     \end{tabular}
  }
\end{table}

\section{Conclusions}
In this work, we have presented Retargetable AR, a novel online AR framework that produces AR experiences facilitating natural interaction between virtual and real worlds according to the pre-set context of AR contents.
To express the scene context, the Retargetable AR generates a 3D semantic map in which geometric and semantic information densely reconstructed and labeled from RGB-D data scanned in the scene is integrated, and constructs a 3D scene graph based on spatial relations between objects in an online manner.
Then it compares the context-representing graph with the AR scene graph corresponding to the context of the AR content as the 3D scene graph is updated and obtains the correspondences between graph nodes, achieving context-aware content arrangements in various scenes. 
The performance and effectiveness of the Retargetable AR were extensively verified through quantitative evaluations of the semantic map and its predicted 3D OBBs, subjective evaluation of the content arrangement results, and online context-aware AR demonstrations in real environments. 
To the best of our knowledge, the proposed approach is the first that considers the semantic registration between the virtual world and the physical world via dense semantic scene understanding, providing an AR experience with natural and realistic interactions.
% この論文では，現実シーンにおいて事前に設定されたコンテキストに応じて，仮想と現実が自然にインタラクションを行うAR体験を実現するためのARフレームワークを提案した．本フレームワークはシーンのコンテキストを表現するために，シーンから取得したRGB-Dデータから幾何的情報と意味的情報を統合した意味的情報マップを取得し，物体間の相対的な空間的関係に基づく3Dシーングラフをオンラインで構築する．この現実シーンのコンテキストを表現したグラフとARコンテンツの想定するコンテキストを表現したグラフとの構造比較を随時行うことで，ノード間の対応関係を得ることでコンテキストを考慮した様々なコンテンツ配置を達成した．その性能と有効性は意味的情報マップと抽出される3次元有向境界ボックスに関する定量評価とコンテンツ配置結果に対する主観評価実験，実環境におけるコンテキストを考慮したARアプリケーションのデモにより示された．本研究は密な意味的シーン理解を通じて仮想と現実の意味的整合性を考慮し，高度で自然なインタラクションを提供するAR技術の先駆けとなるものである．

Finally, we discuss the limitations and open challenges that the proposed framework did not address or include.
In our work, though we present context-aware AR in static scenes, it is desirable to support AR in dynamic scenes and to use the framework in actual environments.
We believe that updating the semantic map while keeping its online performance through the detection of moving objects and relocalization, or object tracking, etc. may be important.
% 最後に，本論文で提案した手法の制約や更なる研究が必要な領域について述べる．
% 本論文では静的環境に対するARコンテンツ配置を示したが，実用する上では動的環境へ対応できることが望ましい．
% そのためには，移動物体の検知や再同定或いは物体のトラッキング等により意味的情報マップをオンライン性能を維持しながら動的に更新することが重要になると考えられる．

The Retargetable AR employed geometric heuristics to estimate the orientations of objects.
However, it is too difficult to accurately estimate them by using heuristics because the objects have a wide variety of geometries, even those of the same category. 
We showed examples in which the combination of semantic information, such as affordance, and the geometric information enabled accurate orientation prediction (e.g., \textit{chair, sofa}).
As another approach to address this problem, we believe that further improvement of a category-level 6DoF object pose estimation method\cite{wang2019normalized} using deep neural networks could contribute to solving this problem. 
Moreover, it is necessary to explore reasonable ways of disentangling the ambiguity of the orientation of symmetric objects and their directional relationships. (E.g., what is a suitable way to define or estimate the orientations of L-shaped sofas and round tables, and what kind of directional relations should be used for expressing their contexts?)
% 本フレームワークでは物体の正面方向を幾何的なヒューリスティックスで推定するに留まっている．
% 物体の形状は多様であり，同一カテゴリ内でのバリエーションも豊かであるため幾何的なヒューリスティックスであらゆる物体の方向を高精度に推定するのは困難である．
% 我々はカテゴリに加えてアフォーダンスのような意味的情報を幾何的な情報と組み合わせて用いることで高精度な推定ができる例を示したが，他のアプローチの1つとしてDNNベースでカテゴリ単位の6DoF物体姿勢を推定する手法\cite{wang2019normalized}の更なる発展がこの課題の解決に寄与すると考えている．加えて，対称物体の向きやそれに起因する空間的関係の曖昧さを合理的に解決する手法も探索される必要がある（例えば，L字型ソファやラウンドテーブルの向きはどう定義・推定され，どういう方向関係を利用するのが妥当か,
% など）．

How to create an AR application with an AR scene graph in a more efficient way or how to scale the graph definition to express various kinds of contexts are also major issues to be addressed.
In this paper, we recognized affordance as an additional attribute of an object, extracted nine kinds of spatial relations, and then defined AR scene graphs by hand based on them to create AR applications.
However, the introduced spatial relations, of course, have limited ability to express contexts, and thus more research is needed to explore more suitable relations, attributes and representations for reflecting the intention of the creators in AR applications more accurately and variously. In addition, the development of authoring tools that will enable us to easily and efficiently generate or develop AR applications that use such attributes or relations for representing contexts is desirable in the future.
% ARコンテンツのコンテキストを反映したARシーングラフを有するARアプリケーションの表現能力の向上や制作効率化も大きな課題の一つである．本論文ではアフォーダンス認識に加えて9つの空間的関係を定義し，これらに基づいてコンテキストを表すARシーングラフを手動で定義してARアプリケーションを制作した．一方で，本論文で定義した関係に基づくコンテキストの表現能力にも限界があるため，ARアプリケーションにコンテンツ制作者の意図をより多様に正確に反映するために必要な関係や属性及びそれらの表現手法は更なる追究が必要である．その上で，そのような属性や関係に基づくコンテキスト表現に基づくARアプリケーションを手動ではなく効率的に生成・作成することができるオーサリングツールの開発が望まれる．

% The authors wish to thank A, B, and C. This work was supported in part by
% a grant from XYZ.}

\bibliographystyle{latex_template/abbrv-doi}

\bibliography{paper}
\end{document}